# Multi-**g**enerational **l**abour **m**arkets: **d**ata-**d**riven **d**iscovery of **m**ulti-**p**erspective **s**ystem **p**arameters using **m**achine **l**earning

Abeer Abdullah Alaql [1], Fahad Alqurashi [1], Rashid Mehmood [2,*]

[1] Department of Computer Science, Faculty of Computing and Information Technology, King Abdulaziz University, Jeddah 21589, Saudi Arabia; aalialaql@stu.kau.edu.sa, Fahad@kau.edu.sa
[2] High Performance Computing Center, King Abdulaziz University, Jeddah 21589, Saudi Arabia
[*] Correspondence: RMehmood@kau.edu.sa

**Abstract:** The current aggressive capitalist approaches have affected social, economic, and planet sustainability. Economic issues, such as inflation, energy costs, taxes, and interest rates, are a constant presence in our daily lives and have been exacerbated by global events such as pandemics, environmental disasters, and wars. A sustained history of financial crises (the 1970s, 2008, FTX, etc.) reveals significant weaknesses and vulnerabilities in the foundations of modern economies. Another significant issue currently is people quitting their jobs in large numbers (Great Attrition). Moreover, many organizations have a diverse workforce comprising multiple generations (e.g., Generations X, Y, Z, alpha) posing new challenges. Transformative approaches in economics and labour markets are needed to protect our societies, economies, and planet. In this work, we use big data and machine learning methods to discover multi-perspective parameters for multi-generational labour markets. The parameters for the academic perspective are discovered using 35,000 article abstracts from the Web of Science for the period 1958-2022 and for the professionals' perspective using 57,000 LinkedIn posts from 2022. We discover a total of 28 parameters and categorised them into 5 macro-parameters, Learning & Skills, Employment Sectors, Consumer Industries, Learning & Employment Issues, and Generations-specific Issues. A complete machine learning software tool is developed for data-driven parameter discovery. A variety of quantitative and visualisation methods are applied and multiple taxonomies are extracted to explore multi-generational labour markets. A knowledge structure and literature review of multi-generational labour markets using over 100 research articles is provided. It is expected that this work will enhance the theory and practice of AI-based methods for knowledge discovery and system parameter discovery to develop autonomous capabilities and systems and promote novel approaches to labour economics and markets, leading to the development of sustainable societies and economies.

## 1. Introduction

### 1.1 Global Economic Challenges

Economic issues, such as inflation, energy costs, taxes, and interest rates, are a constant presence in our daily lives. These issues have been exacerbated by global events such as the COVID-19 pandemic, environmental disasters, geopolitical tensions, and wars. These issues are a source of concern for experts, the media, politicians, and ordinary citizens alike. For instance, in the US, , the ongoing COVID-19 pandemic and its variants and vaccination efforts, labor shortages, supply chain vulnerabilities, the policies of the Federal Reserve, and US-China relations have been identified as some of the top economic risks of 2022 [1]. Global inflation, low consumer spending, the impact of climate change on global economies, rising labor costs, gas supplies for Europe, and global food security pose major concerns for the global economy [2].

Our societies and workplaces are becoming more diverse and divided, with increased separation along various dimensions such as age, gender, race, and ideology. This has resulted in a greater sense of polarization and conflict within our communities, even in areas such as sports where traditionally the purpose was to create love and harmony among people. This level of complexity and division is relatively new and not something that we have traditionally experienced to the same extent.

### 1.2 Great Attrition

To add to those economic challenges, there is currently a significant issue of people quitting their jobs in large numbers, known as the Big Quit, Great Attrition, or Great Resignation, which has been accompanied by the phenomenon of Quiet Quitting [3,4]. This trend has been observed globally, with many workers citing factors such as a desire for flexible work arrangements, a disconnect between business leaders and employees, high productivity expectations leading to burnout, and difficulties with work-life balance as reasons for considering a job change [5,6]. Key factors that influence the decision to change jobs include compensation, a sense of meaning and self-fulfillment in the work, confidence and competence in the role, and autonomy in terms of time, place, and flexibility. The concept of Quiet Quitting is viewed differently by different people, with some seeing it as a way to do the minimum at work or to separate careers from identities, and others seeing it as a reminder to avoid burnout [3].

### 1.3 Multi-Generational Workforce (Age Dynamics)

Many organizations face the unavoidable consequences of having a diverse workforce. There is a strong correlation between this phenomenon and the presence of multiple generations in the workplace, which is one of the dynamics that has a tremendous effect on the composition of the workforce. Generations are a popular term used to describe the demographic groupings that occur within a population over the course of their lives. Each generation is a group of people who share a particular set of experiences, outlooks, and values due to the environment in which they were raised. Research has shown that employees tend to display certain tendencies towards work, which are dictated by their generational experiences. Several other factors contribute to these differing work attitudes, including technology, economic status, gender, education, and ethnicity. The impact of generational differences and other factors can be seen throughout the workplace, and it is important for organizations to pay attention to these factors so that their workforce can be productive and engaged.

Well-known generations based on the time of their birth are Alpha Generation, Generation Z, Millennials (also called Generation Y), Generation X, Baby Boomers, Silent Generation, and Greatest Generation (see Section 4.3 for an introduction to these generations). Generations (first, second, and third generations) could also refer to generations of immigrants in the US or other countries.

### 1.4 Research in Multi-Generational Labour Markets

A sustained history of financial crises such as the financial crises of the 1970s and 2008, and the recent FTX collapse reveal significant weaknesses and vulnerabilities in the foundations of modern economies, including the financial system, regulatory frameworks, and economic policies. Moreover, the current aggressive capitalist approaches have affected social and planet sustainability. Transformative approaches in economics are needed to protect our societies, economies, and planet.

Labour markets are an important part of the economy that have significant impacts on people, businesses, and society. They play a crucial role in determining employment and wages, and they are a key factor in economic growth. Labour markets can also contribute to income inequality, and they can affect people's quality of life. Studying labour markets can help us understand how these factors work together and how different economic policies and conditions can impact employment, wages, and overall economic outcomes. Research in labour markets can help us better understand the changing nature of labour and employers, as well as the role of society in labour markets. The research is important also to incorporate national and global priorities, such as the UN Sustainable Development Goals while developing labour economics and markets. These considerations would allow the development of sustainable cities and societies.

Studying generations in labour markets allows for an understanding of different age groups' experiences and challenges, aiding informed decisions for supporting and addressing their needs. It helps to understand younger workers' challenges, such as student debt and competitiveness, and older workers' challenges, such as ageism and technology adaptation. This understanding creates a more inclusive and supportive work environment for all.

### 1.5 This work

In this work, we use big data, machine learning, web scraping, and other cutting-edge methods to discover multi-perspective parameters for multi-generational labour markets. The two different perspectives of labour markets are provided by two types of data sources: one seen by academics and researchers (using academic literature) and the other by industry, employers, and professionals (using Linked). Despite differing perspectives with considerable differences, these views are mutually inclusive and complement one another.

Our academic-view dataset, derived from the Web of Science, an academic database, allows us to identify parameters for academia-focused aspects of labour markers. We collected around 35K article abstracts (including titles and keywords) for the period 1958-2022 from different categories of academic disciplines on the Web of Science, for example, Educational Research, Business Economics, Computer Science, Social Sciences, Demography, Communication, and Others. We discovered 15 parameters from the academic dataset using the Latent Dirichlet Allocation (LDA) algorithm and categorized them into five macro-parameters namely

Learning & Skills, Employment Sectors, Consumer Industries, Learning & Employment Issues, and Generations-specific Issues.

The LinkedIn dataset comprises 57K posts collected from LinkedIn using web scraping. We discovered 13 parameters from the data using LDA. The macro-parameters are the same as for the Web of Science data, however, there are differences in their constituent parameters. In this paper, we only give the names of the parameters and macro-parameters discovered from LinkedIn and use them to develop the multi-perspective taxonomy of multi-generational labour markets. A detailed description of the parameters can be found in [7].

We implement a complete software tool to discover parameters. This tool consists of four software components: data collection and storage, pre-processing, parameter modeling, and visualization and validation. Both datasets are pre-processed using web scraping and other techniques to remove duplicate and irrelevant data, tokenize data, clean up the data, and lemmatize data to prepare them for LDA analysis.

The document clusters and domain knowledge are used to discover each dataset's labour market parameters and macro-parameters. A wide range of quantitative analysis methods is applied to the clustered data, including term scores, keyword scores, and intertopic distance maps. Different visualization methods describe datasets, document clusters, and discovered parameters. Word clouds, taxonomies, and dataset histograms are some examples. Multiple taxonomies of the labour market are extracted from the perspectives of academics, employers, and employees. According to professional platforms and academia, Figure 1 illustrates a high-level multi-perspective taxonomy of the labour market. As shown in the figure, the first-level branches represent macro-parameters and the second-level branches represent parameters. A validation of our results can take place internally or externally. Further details of validation and other aspects of the methodology and the software tool are provided in Section 3.

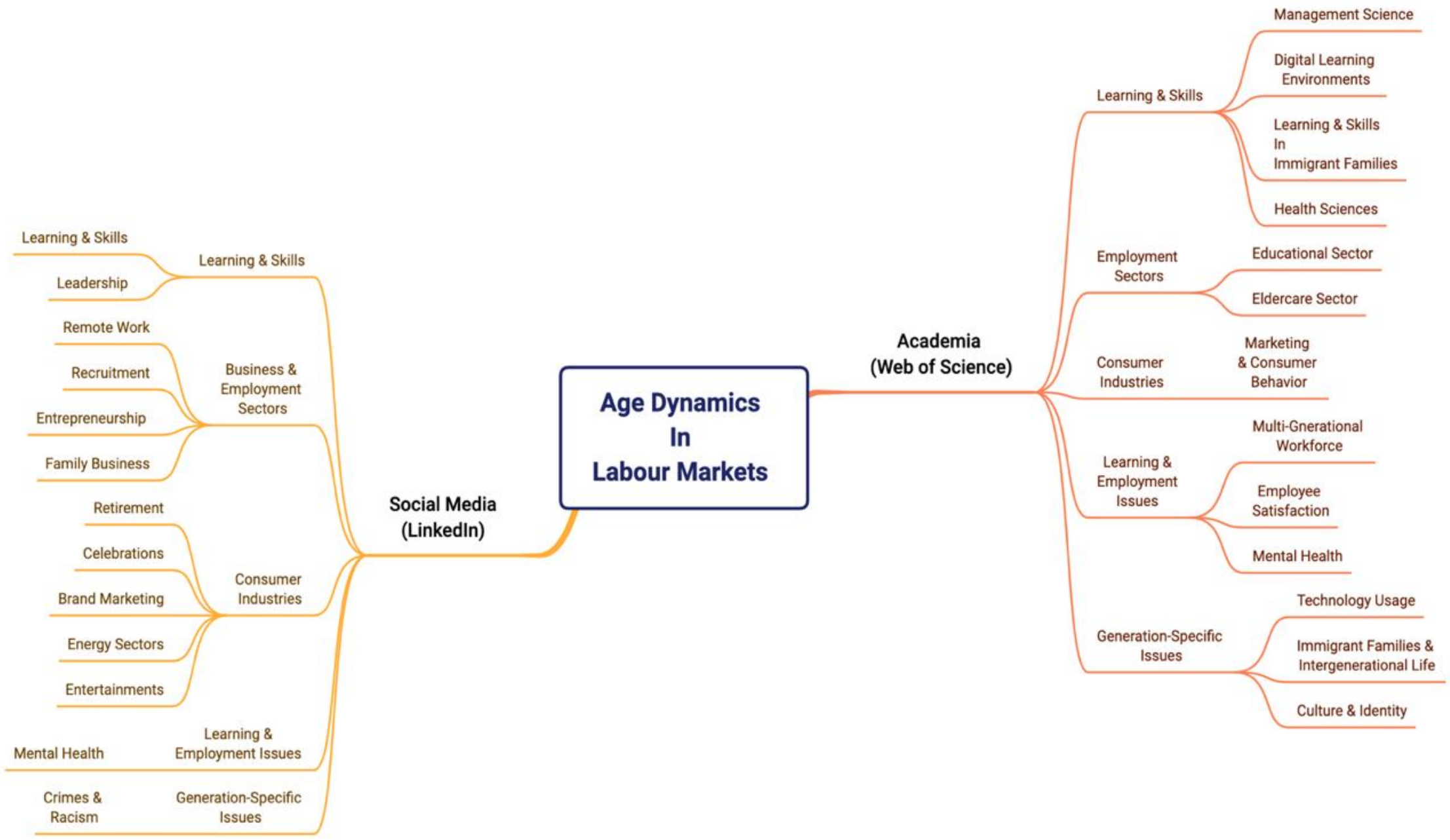

Figure 1 A multi-perspective taxonomy of Multi-Generational Labour Markets

The paper also provides a knowledge structure and literature review of multi-generational labour markets using over 100 research articles.

### 1.6 Utilisation and Novelty

This paper contributes an approach that makes it possible to obtain comprehensive, objective, and multi-perspective information on a subject using machine learning and deep learning. It also provides tools and resources that allow anyone to access information about matters of public interest from various datasets. The research in this paper contributes to our understanding of age dynamics in labour markets and may be used to raise awareness of important issues among the public, academics, and others, as well as to drive future research on this topic using advanced technologies. The findings and knowledge gained from this work can be used by the public to make informed decisions and can guide labour economics research in important areas. It is expected that this work will enhance the theory and practice of AI-based methods for information discovery, and system parameter discovery to develop autonomous capabilities and systems, extend the use of LinkedIn and scientific literature media for information and parameter discovery, and promote novel approaches to labour economics and labour markets through the knowledge discovery approach, ultimately leading to the development of sustainable societies and economies.

Artificial intelligence is increasingly allowing autonomous functionality in systems such as self-driving cars, web services, drones, and robots in manufacturing and farming. The autonomous capability will increasingly become part of all systems around us and will be extended to larger systems including industrial sectors, city, and country governance. Even in the absence of autonomous capabilities discovering system parameters is necessary because they provide a foundation for decision-making and problem-solving during the design and operations process.

We will see in Section 2 (Related Works) that the previous analyses of LinkedIn data by machine learning focused primarily on profile information to understand skills, jobs, and relationships. Meanwhile, earlier machine learning-based analysis of the Web of Science has focused on other topics. No previous work has automated the detection of multigenerational perspectives (parameters) for age dynamics in labour markets based on LinkedIn posts, academic literature from the Web of Science, or both. We have made novel contributions in the area of age dynamics in labour markets, the methodology of discovering parameters for labour markets, and the quantitative and descriptive analysis of age dynamics in labour markets. We analyse multi-generational labour markets in this study using data from academic literature on the Web of Science and social media posts on LinkedIn, in contrast to the traditional method of studying multi-generations in psychology, which involves asking a small group of people to answer survey questions about a group of interest. We can gain a deeper understanding of multi-generational interactions in social interactions and academic literature by investigating how they form and spread. To intervene with educational, counter-narrative, and other mitigating measures, it is necessary to identify and monitor the age dynamics of currently pervasive group perceptions.

In this paper, the sections are arranged as follows. The related works and research gap are presented in Section 2. The approach, the methodology, and the tools are described in Section 3. Sections 4 and 5 discuss the parameters discovered for Web of Science and LinkedIn, respectively. Section 5 includes a discussion. Conclusions and future directions are presented in Section 6.

## 2. Related Works

In this section, we discuss the work related to our paper. A comprehensive review of literature on using machine learning-based analytics for various applications was conducted. No work directly related to our work was found. Thus, we review two work areas to situate our work within the context of data analytics in a multi-generational workforce. We begin with a literature review on the use of natural language processing (NLP) in scientific literature. Following that, we examine the use of artificial intelligence and machine learning across related fields on social media including LinkedIn.

### 2.1 NLP on Academic Literature

Porter et al. [8] aimed to obtain trends and tools in the DarkNetMarkets subreddit, along with the changes in those tools and trends, through generating topic models using LDA. The subreddit posts were taken as the input data.

Maier et al. [9] reviewed the research literature dealing with significant challenges that needed to be tackled for applying LDA to textual data and proposed a methodology to address these challenges. A user guide for applying LDA for topic modeling was also developed. Kherwa et al. [10] presented a study on topic modeling after analysis of around 300 research articles, including topic modeling methods, Posterior Inference techniques, classification hierarchy, and different evolution models of LDA and their applications in various technological domains. A quantitative evaluation of topic modeling methods and a comprehensive discussion on the challenges of topic modeling were also presented. In addition, Russo et al. [11] investigated the nature of the research conducted on personality in the clinical field using LDA. Yun et al. [12] proposed another approach for automatic patent classification by using LDA for text representation and its results as an input for a support vector machine (SVM). Bastani et al. [13] proposed a decision support system for consumer complaint analysis and used topic modeling to discover consumer issues automatically. Chen et al. [14] used topic modeling and sentiment feature analysis to reveal considerable differences between the use of controversial and non-controversial terms related to COVID, such as “Chinese virus” and “COVID19,” respectively. Dong et al. [15] extracted Corona Virus-related abstracts, used LDA to train an eight-topic model from the corpus, and compared the topic distribution between other Corona Virus infections and COVID-19. Selvi et al. [16] proposed a temporal rule-based classification algorithm by integrating LDA and Naïve Bayes Classifier for efficient storage and speedy retrieval of medical data.

### 2.1 NLP on Social Media and LinkedIn

A social media data mining process involves gathering and analyzing social media data. Data mining aims to find patterns in social media data using machine learning and other advanced technologies. For instance, it can be used to study the demographics of its users [17]. Companies can also use it to identify employee skills and qualifications. Alruwaili et al. [18], for example, extracted LinkedIn profiles and used clustering techniques based on their skills. Likewise, Dai et al. [19] classified educational and cluster professional backgrounds. A study by Purwono et al. [20] evaluated the power of LinkedIn profile photos to predict face shapes. Nguyen et al. [21] also explored the LinkedIn profiles and facial features. Rsearchers tested the feasibility of using IT professionals' data as a reference source [22], [23]. Also, Domeniconi et al. [24] discovered relationships between jobs and people skills based on LinkedIn user profiles. Giri et al. [25] used LinkedIn data to allow recruiters to select. Based on Greedy, Hierarchical, and K Mean clustering algorithms, Garg et al. [26] clustered LinkedIn profiles based on their job title, company name, and geographic locations. Similarly, Li et al. [27] applied deep transfer learning to develop domain-specific job understanding models.

Machine learning has also been used in various research fields to extract information from other types of social media data, such as Twitter data. According to Haghighi et al. [28], transit riders' perceptions of service quality can be evaluated using Twitter data. The authors use unsupervised machine learning to filter tweets related to the transit system's actual customer experiences to generate topic models. The opinions of transit riders are then evaluated using an established tweet-per-topic index to discover the reasons for their unhappiness. This methodology may be helpful to transportation authorities for user-oriented analysis and investment decisions. A further approach was developed by Alomari et al. [29] to identify traffic-related events based on machine-learning algorithms based on Twitter data in Arabic. Additionally, Karami et al. [30] summarized the temporal trend of Twitter-based studies over the last decade and interpreted their evolution. Moreover, Martínez-Rojas [31] et al. provided an overview of the current state of research on the use of Twitter in emergency management and identified problems and potential suggestions for further studies. There are many other works on the use of Twitter data such as [32–35].

### 2.3 Research Gap

The earlier machine learning-based analysis of the Web of Science has focused on other topics. The previous analyses of LinkedIn data using NLP focused primarily on profile information to understand skills, jobs, and relationships. No previous work has automated the detection of multigenerational perspectives (parameters) for age dynamics in labor markets based on LinkedIn posts, academic literature from the Web of Science, or both. We have made novel contributions in the area of age dynamics in labor markets, the methodology of discovering parameters for labor markets, and the quantitative and descriptive analysis of age dynamics in labor markets.

## 3. Methodology & Software Tool Design

This section explains the methodology and design of the proposed architecture for the system. In Figure 2, we illustrate the architecture of our software, which is composed of four components, each of which is analyzed in more detail below. A description of the methodology is provided in Section 3.1, which also includes the master algorithm. Data collection, data sources (Web of Science and LinkedIn[7]), preprocessing, parameter modeling, parameter discovery and quantitative analysis, validation, and visualization are explained in sections 3.2 to 3.9.

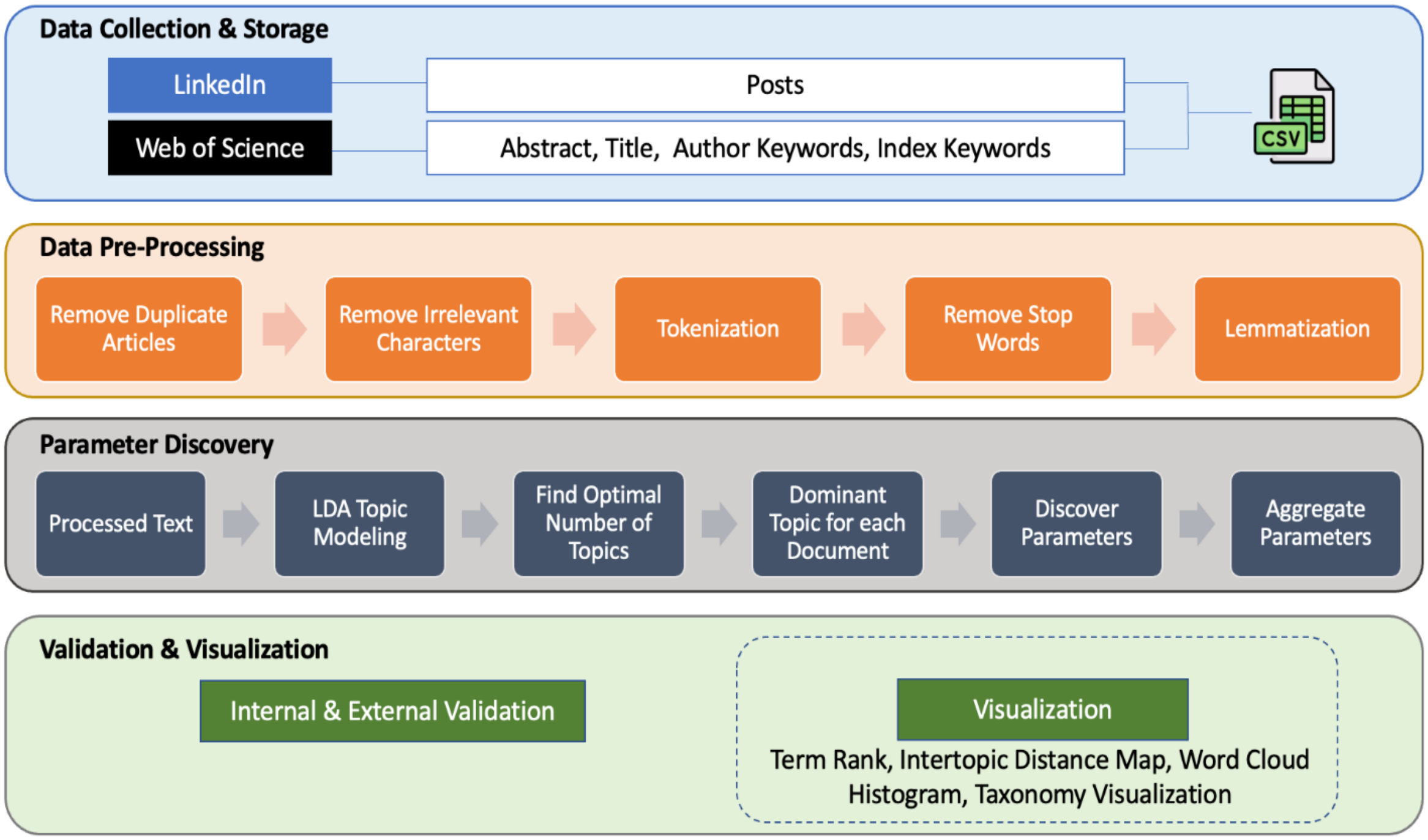

Figure 2 The system Architecture

### 3.1 Methodology Overview

First, we introduce Algorithm 1, the master algorithm of our system. The algorithm has been called two times since we have two datasets. For the first dataset, we download the CSV file directly from Web of Science, which includes Article Titles, Abstracts, Keywords, and publish date. On the other hand, for the second dataset, we collected LinkedIn posts using web scraping techniques (e.g., Python, Beautiful Soup, Requests, and Pandas), as shown in Algorithm 2. The post objects were collected as JSON objects (JavaScript Object Notation). Afterward, we converted the JSON file to a CSV file.

Algorithm 1 takes the CSV file as input. Then, the CSV file is preprocessed in the following step using Pandas. Next, we used Gensim's Python library for parameter discovery using LDA[36]. Our quantitative analysis was implemented using domain knowledge, the clusters were classified as parameters, and then the parameters were grouped into macro-parameters. Our final step was to visualize the parameters and macro-parameters. In addition, these parameters were validated internally and externally.

**Algorithm 1 Master Algorithm**

Input: CSV file
Output: Visualization of the discovered parameters

1: CSV_file
2: Data_dataframe ← Read CSV_file
3: Processed_data ← Preprocessing data_dataframe
4: Coherence_values ← Compute coherence for different LDA models
5: Optimal_num_topics ← Find optimal number based on coherence_values
6: Optimal_model ← Load optimal LDA model
7: LDA_visualization ← Apply LDAvis to optimal_model

**Algorithm 2 Data Collection (LinkedIn)**

```
Input: search_query
Output: CSV_file

Link ← Load linkedin weblink
Driver ← Load chromedriver
Login_field ← Login to linkedin
Search_query ← Search_query
Driver ← Load posts page for search_query
Last_height ← Get the recent height of the posts page
while height of page keeps changing
        New_height ← get the new height of the page
        if New_height is equals to Last_height then
                begin
                        exception
                                break the loop
                end
        end if
        Last_height ← New_height
end while
Src ← Load posts page source code
Post_details ← extract description of all posts from the Src
Post_dict ← save post content in a dictionary from Post_details
JSON ← dump Post_dict
JSON_file ← read JSON
CSV_file ← create CSV from JSON_file
```

*3.2 Data collection*

We conducted an extensive literature review to identify alternative synonyms for a multi-generational workforce.

*3.3 Dataset: Academic literature (Web of Science)*

We used the Web of Science database to obtain the most relevant documents with an integrated query language and data format. Moreover, it promotes access to topic indexes, citation indexes, and other databases from other disciplines, which can be used to find relevant research and evaluate its conclusions. Thus, we have collected our dataset from the Web of science. The dataset was generated using a Boolean query (Q1), as shown in Table 2. We have collected around 35K articles (1958-2022) from several Web of Science disciplines, for example, Educational Research, Business Economics, Computer Science, Social Sciences, Demography, Communication, and Others. The number of articles after removing duplication is around 31K.

*3.4 Dataset: Posts(LinkedIn)*

In our previous study[7], we have collected our dataset from LinkedIn (posts) for four months (March 2022 – July 2022) using search query terms shown in Q2, Table 1. These posts are anonymized (aggregated) data, and we do not use personal information. Then, posts are saved in a CSV file, including around 57K posts. The CSV file includes two columns, the query term, and the post.

**Table 1.** Data Collection

| **Source** | **Query number** | **Query** | **Data size** |
|---|---|---|---|
| Web of Science | Q1 | "(First generation OR second generation OR Y-ers OR X-ers OR Millennial* OR Gen Y OR Generation Y OR Generation X OR Gen X OR Gen Z OR Generation Z OR baby boomer* OR multi-generation OR generation* OR Inter-generational OR cross-generational OR cohort OR demographic*) AND (workforce OR labour force OR workplace OR skill* OR Qualification* OR experience* OR work OR employment OR occupation)" | 35K |
| LinkedIn | Q2 | "Millennials, Gen Y, baby boomers, Generation Y, Generation X, Gen X, Gen Z, Generation Z, Alpha generation, multi-generation, Inter-generational, cross-generational, Employment, Unemployment, Middle Aged, Aged, Elderly, Young, Senior, Seniors, 20s, 30s, 40s, 50s, 60s, 70s, 80s, 90s, Ageism, Labourforce, Demographics, Workforce, Workplace, Workers, Retirement, Recruit, Aging, Leadership, Entrepreneur, Preschool Child, Child, Preschool, Infant, Children" | 57K |

3.5 Data Preprocessing

Preprocessing steps, as shown in algorithm 3, include removing duplicate articles and irrelevant characters, tokenizing, stopping words, and lemmatizing with POS tags. As a first step, Python's Pandas package reads and saves the CSV file as a data frame (DF). As a second step, we removed all redundant data. The number of articles after removing duplication is around 31K. Third step, we eliminated all unnecessary characters, including several Unicode characters. The fourth step of the process was to tokenize the texts using a simple preprocess function from the Python package "Gensim.". In the fifth step, stop words were deleted from the article. The final step of preprocessing the data was lemmatization using Spacy. As a result of preprocessing, the cleaned texts were obtained, which were then used in the LDA model.

**Algorithm 3 data-pre-processing**

```
Input: CSV_file
Output: processed_data

1: data_df ← read CSV_file
2: removed_dup_df ← remove duplicate data_df
3: removed_irr_char_df ← remove irrelevant characters from removed_dup_df
4: token_df ← tokenize removed_irr_char_df
5: rsw_df ← remove stopwords from token_df
6: lemm_df ← lemmatize rsw_df
7: processed_data← clean  lemm_df
```

3.6 Topic Modeling Using Unsupervised Machine Learning

A topic model is a machine learning technique for discovering topics in a set of documents [37]. Information filtering, retrieval, and semantic search are some applications of topic modeling used in document analysis. A topic model can also be referred to as a latent topic model or a latent semantic model. This method analyzes words and their variations in various contexts to discover hidden themes in a text.

We used Latent Dirichlet Analysis (LDA), a topic modeling method, in this study. We set the parameters for our LDA model to 15 topics, 10 passes, and 100 iterations. There is no doubt that the number of topics is an important parameter when building a model. When the number of topics is large, the model must be overfitted, but when it is small, it must be underfitted. By creating multiple LDA models with different topics (k) and comparing the coherence measures with the visual representation, we could determine the optimal number of topics. Passes refer to the number of times an algorithm must traverse the whole corpus. The maximum number of iterations is required to determine the probability of each topic in the corpus.

### 3.7 Coherence Measures

In topic modeling, coherence scores are used to measure the human interpretability of a topic. The topic coherence scores give an extensive method for comparing various topic models. It compares by capturing the optimal number of topics and gives a 'Coherence Score' for those topics' interpretability. Recent research has focused on measuring the coherence of topics to address the issue that topic models do not guarantee the interpretability of their output. The most common methods for adjusting LDA hyperparameters are based on various topic coherence measures. In this work, we used the C_v, C_uci, C_umass, and C_npmi measures of coherence. C_v measure uses a sliding window-based one-set segmentation of the most popular or top words and an indirect confirmation measure. The indirect confirmation measure employs cosine similarity and normalized pointwise mutual information (NPMI). C_uci measure employs a sliding window and all word pairs' pointwise mutual information (PMI). C_umass measure is a confirmation measure, and it utilizes co-occurrence counts for documents, a one-preceding segmentation, and a logarithmic conditional probability. C_npmi measure can be considered an enhancement of C_uci coherence, which uses the normalized pointwise mutual information (NPMI).

## 3.8 Parameter Discovery & Quantitative Analysis

We identify the parameters and macro-parameters based on domain knowledge and quantitative analyses utilizing tools like term scores, intertopic distance maps, and word clouds.

### 3.8.1 Term Score

The keyword scores for each parameter are represented graphically in descending order. This term score visualization substantially influences parameter identification.

### 3.8.2 Intertopic Distance Map

A two-dimensional intertopic distance map depicts the parameters as parameter circles whose sizes correlate to the number of words in the dictionary that define the parameters.

### 3.8.3 Word Clouds

A word cloud is a representation of a group of words. Word Cloud highlights the famous words and phrases in the articles based on how often they appear and how important they are. It gives quick, easy-to-understand graphics that can lead to in-depth analyses.

## 3.9 Validation and Visualization

Results can be both internally and externally validated. Internal validation of a parameter means studying and reviewing the relevant papers. In our research, documents may be academic articles or posts. We described how we interpreted the relationship between the papers and the parameters. External validation is performed by comparing the two datasets' parameters, keywords, and metric metrics. For internal and external validation, many visualization methods are utilized for the visualization. Numerous visualization techniques are utilized to describe the datasets, document clusters, and the discovered parameters using histograms, taxonomies, Term scores, Intertopic Distance Map, and word clouds. Several Python packages, such as LDAvis, Plotly, and Matplotlib, are used to construct these visualizations.

## 4. Parameter Discovery for Multi-Generational Labour Markets (Web of Science)

In this section, we describe the parameters that were detected using our LDA model based on the LinkedIn datasets. The parameters are divided into five macro-parameters. The parameters and macro-parameters are discussed in Section 4.1. Section 4.2 gives the quantitative analysis of the parameters. The five macro-parameters are discussed in Sections 4.3 to 4.7.

### 4.1 Overview and Taxonomy

Based on the LDA model, 15 parameters were found in the LinkedIn dataset. This 15-parameter set was grouped into five macro-parameters. Based on the coherence score, we determined 15 clusters (which will be discussed in more detail in the next section). Table 2 lists the LinkedIn dataset parameters and macro-parameters. According to Column 1, the parameters are categorized into five macro-parameters: Generations-specific Issues, Learning & Skills, Employment Sectors, Consumer Industries, and Employment Issues. Columns two and three list parameters and cluster numbers, respectively. According to the fourth column, each parameter has a certain percentage of keywords. As indicated in the fifth column, each parameter is associated with the top keywords.

In Figure 3, we show a taxonomy of the multi-generational workforce. To create the taxonomy, we used the parameters and macro-parameters from the Web of Science dataset. First level branches display macro-parameters, and second level branches display discovered parameters.

**Table 2.** Parameters and Macro-parameters of multi-generational labour markets

| Macro-parameters | Parameters | No | % | Keywords |
|---|---|---|---|---|
| Learning & Skills | Digital Learning Environments | 3 | 9 | System, Data, Model, Generation, Base, Propose, Method, Using, Network, Approach, Work, Problem, Use, Generate, Application, Result, Time, Paper, Analysis, User. |
| Learning & Skills | Digital Learning Environments | 11 | 4.2 | Learning, Student, Education, Course, Teaching, Skill, Experience, Language, Base, Study, Group, Learner, Online, Environment, Cohort, Design, Educational, Classroom, Assessment, University. |
| Learning & Skills | Management Science | 2 | 11.1 | Social, Community, Experience, Work, Research, Practice, Identity, Interview, Study, Within, Article, Cultural, Explore, Context, Generation, Theory, Qualitative, Challenge, Paper, Diversity. |
| Learning & Skills | Health Sciences | 14 | 3.5 | Medical, Training, Study, Skill, Clinical, Program, Result, Method, Survey, Assessment, Practice, Cohort, Evaluation, Professional, Participant, Knowledge, Test, Scale, Group, Question. |
| Learning & Skills | Learning & Employment in Immigrant Families | 10 | 5.1 | Student, Education, University, Academic, Higher, Study, College, School, Program, Graduate, Career, Science, First, Research, Year, Experience, Skill, Faculty, Cohort, Degree. |
| Employment Sectors | Education Sector | 4 | 8.6 | Teacher, School, Education, Teaching, Professional, Educational, Science, Development, Primary, Study, Training, Educator, Practice, Secondary, Service, Covid-19, Experience, Classroom, Pandemic, Curriculum. |
| Employment Sectors | Eldercare Sector | 8 | 7 | Health, Care, Services, Mental, Police, Service, Need, Patient, Support, Trauma, Older, Life, Work, Disability, Social, Treatment, Experience, People, Population, Caregiver. |
| Consumer Industries | Marketing & Consumer Behavior | 15 | 2.1 | Consumer, Customer, Product, Marketing, Service, Brand, Food, Consumption, Experience, Market, Purchase, Value, Industry, Fashion, Tourism, Price, Environmental, Retail, Housing, Quality. |
| Learning & Employment Issues | Multi-Generational Workforce | 13 | 3.6 | Knowledge, Research, Development, Management, Business, Generation, Design, Process, Innovation, Paper, Approach, Project, Model, Base, Work, Study, Need, Value, Practice, Methodology. |
| Learning & Employment Issues | Employee Satisfaction | 7 | 7.9 | Example, Study, Work, Factor, Employee, Satisfaction, Demographic, Relationship, Research, Findings, Result, Performance, Difference, Influence, Gender, Attitude, Variable, Level, Intention, Perception, Analysis. |
| Learning & Employment Issues | Mental Health | 9 | 5.6 | Risk, Violence, Stress, Self, Experience, Study, Factor, Abuse, Adolescent, Among, Social, Sexual, Emotional, Behavior, Relationship, Burnout, Demographic, Health, Report, Associate. |
| Generations-specific Issues | Technology Usage | 1 | 11.5 | Information, Technology, Digital, Medium, Social, Library, Online, User, Communication, Internet, Literacy, Game, Generation, Mobile, Computer, Study, Network, Experience, People, Research. |
| Generations-specific Issues | Culture & Identity | 12 | 3.9 | Generation, Cultural, Culture, Life, Political, Chinese, History, Article, World, Identity, Society, People, Memory, Young, Values, Experience, Story, Narrative, Historical, Language. |
| Generations-specific Issues | Immigrant Families & Intergenerational Life | 6 | 8.5 | Child, Family, Woman, Parent, Cohort, Immigrant, Life, Intergenerational, Study, Mother, Adult, Fertility, Gender, Data, Employment, Experience, Among, Young, Time, Result. |
| Generations-specific Issues | Immigrant Families & Intergenerational Life | 5 | 8.5 | Country, Economic, Policy, Employment, Market, Change, Demographic, Social, Population, Growth, Migration, Financial, State, Increase, Rural, Labour, Economy, Worker, Migrant, Development. |

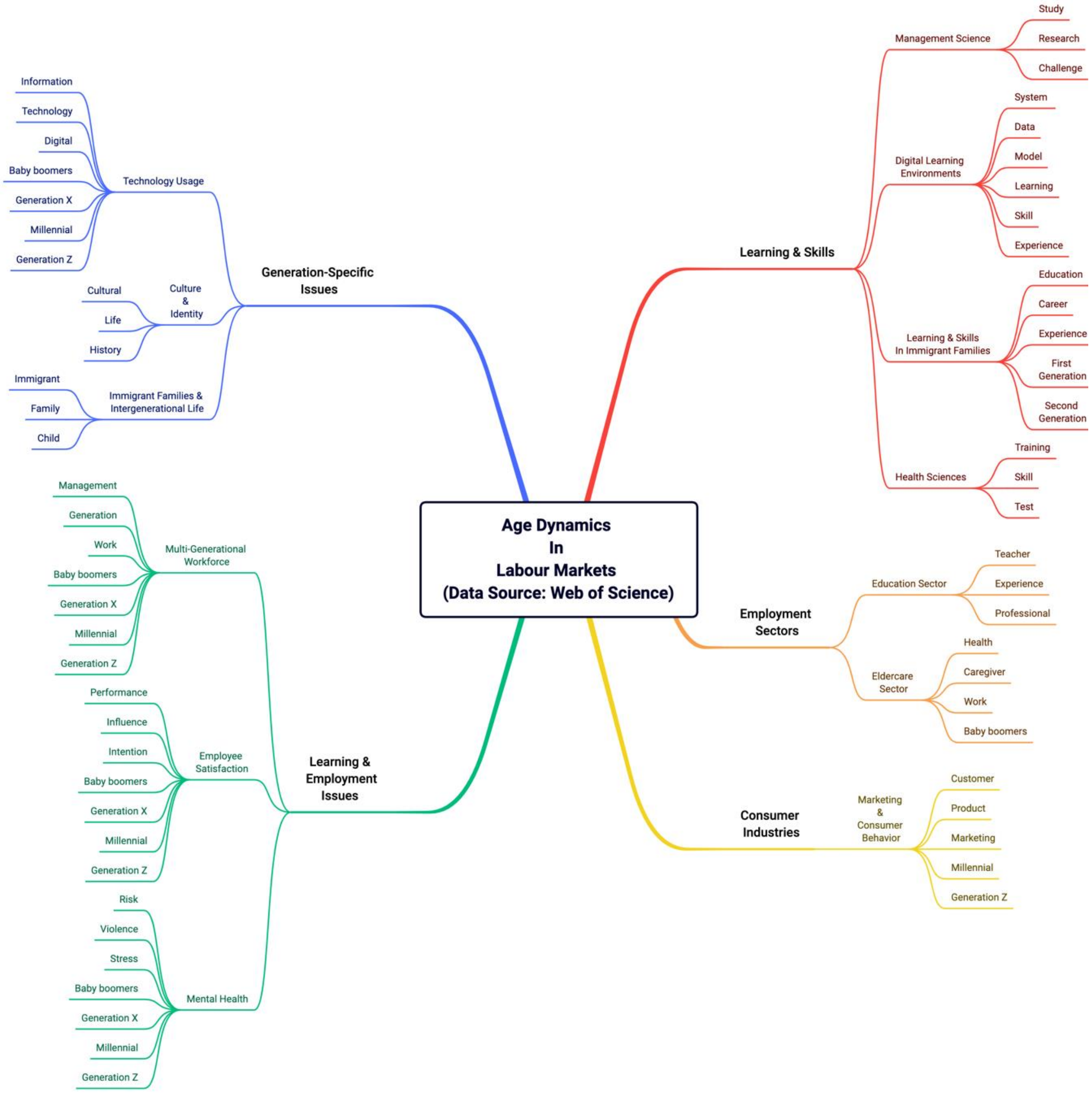

Figure 3 Web of Science perspective taxonomy of Multi-generational Labour Market

4.2 Quantitative Analysis

The coherence measures, intertopic distance map, term score, word clouds and histograms are discussed in this section. Models were created with Gensim LDA package and themes were rendered with PyLDAvis. Based on the coherence measures, we selected different top models with the optimal number of topics. Figure 4 presents four different coherence measures for finding the optimal number of topics. Our choice of 15 topics was since it appeared to be an optimal number. Furthermore, the intertopic distance map has fewer overlapping circles.

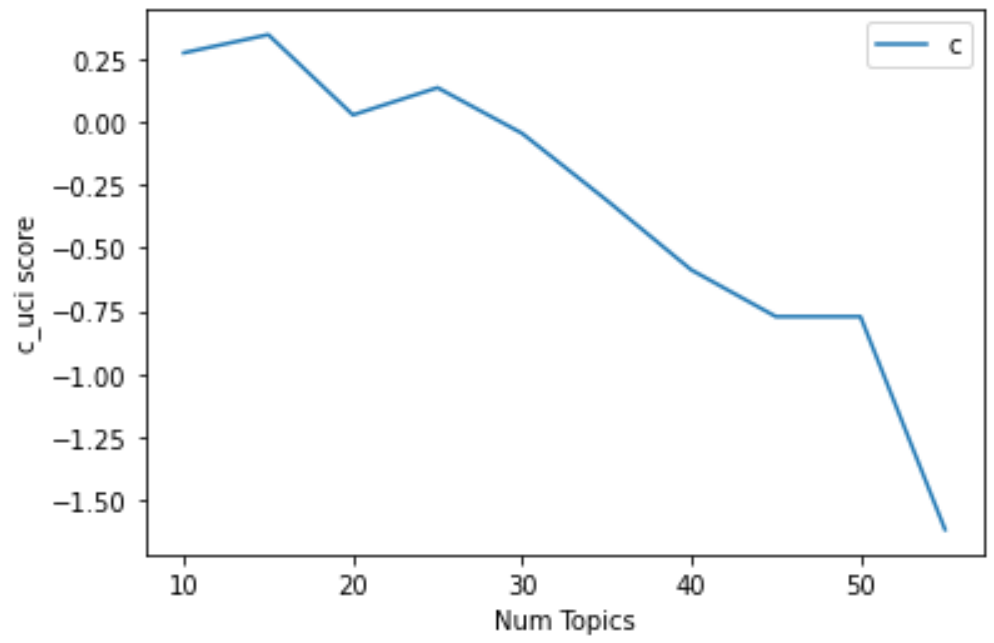

(a)_V Score

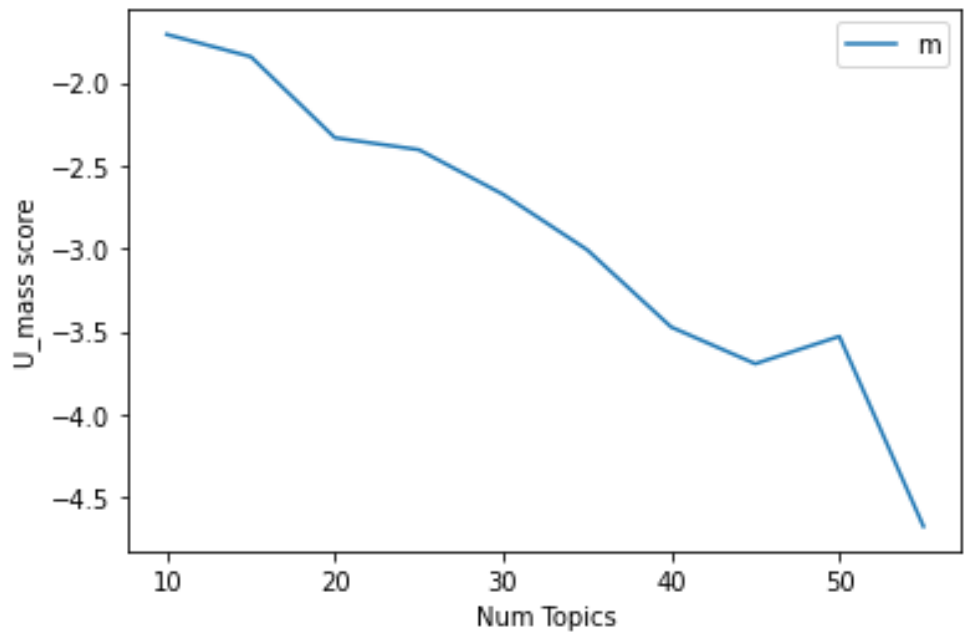

(b) U_mass Score

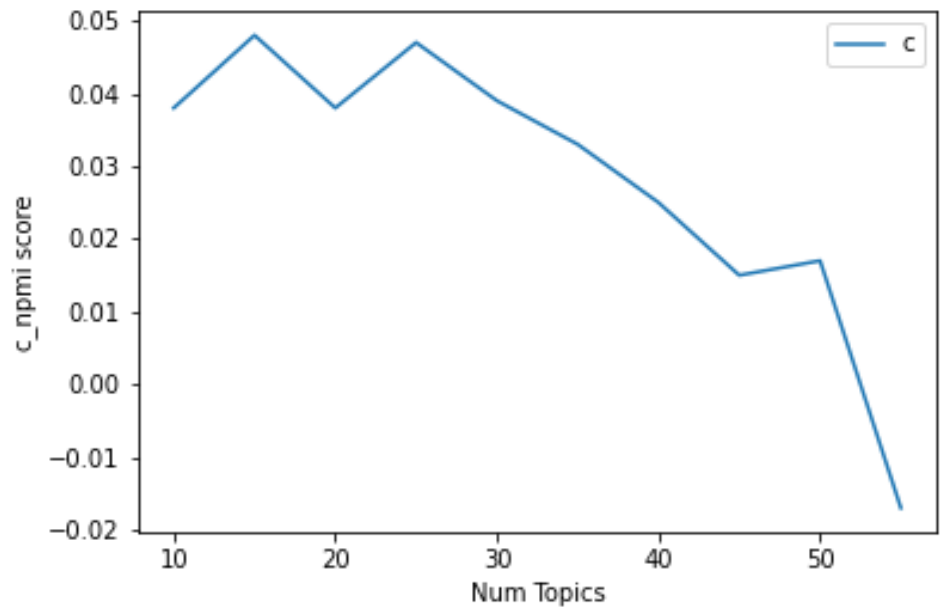

(c) C_uci Score

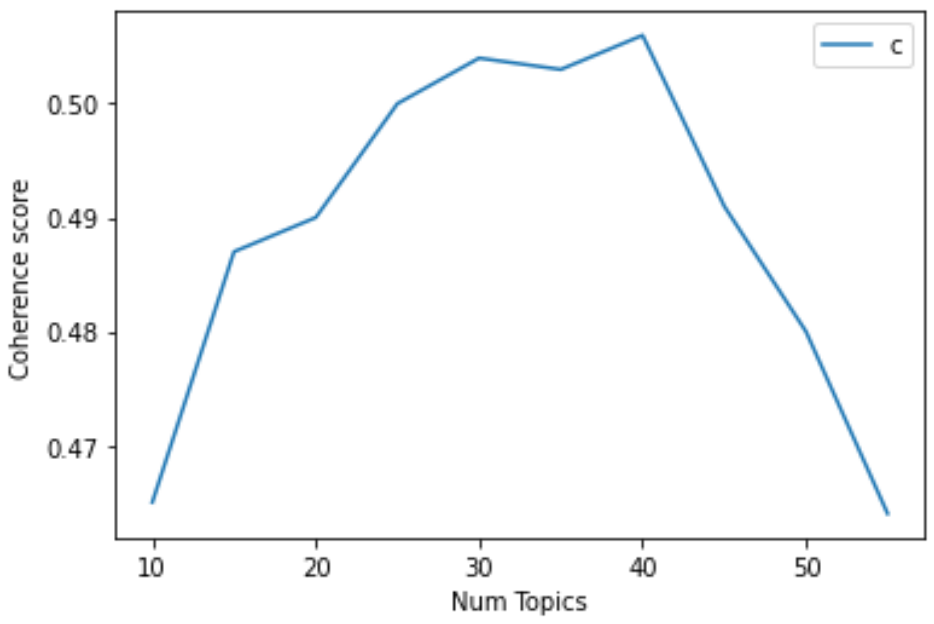

(d) C_npmi Score

Figure 4 Coherence measures

Based on the extracted 15 topics information, Figure 5 shows the inter-topic distances and the most important words. Topics are represented by circles on the intertopic distance map. Depending on the topic's relevance, the circle size increases significantly. Typically, a good topic model will have fewer overlapping circles throughout the chart. A poor topic model, on the other hand, would have circles that overlap heavily clustered in one quadrant.

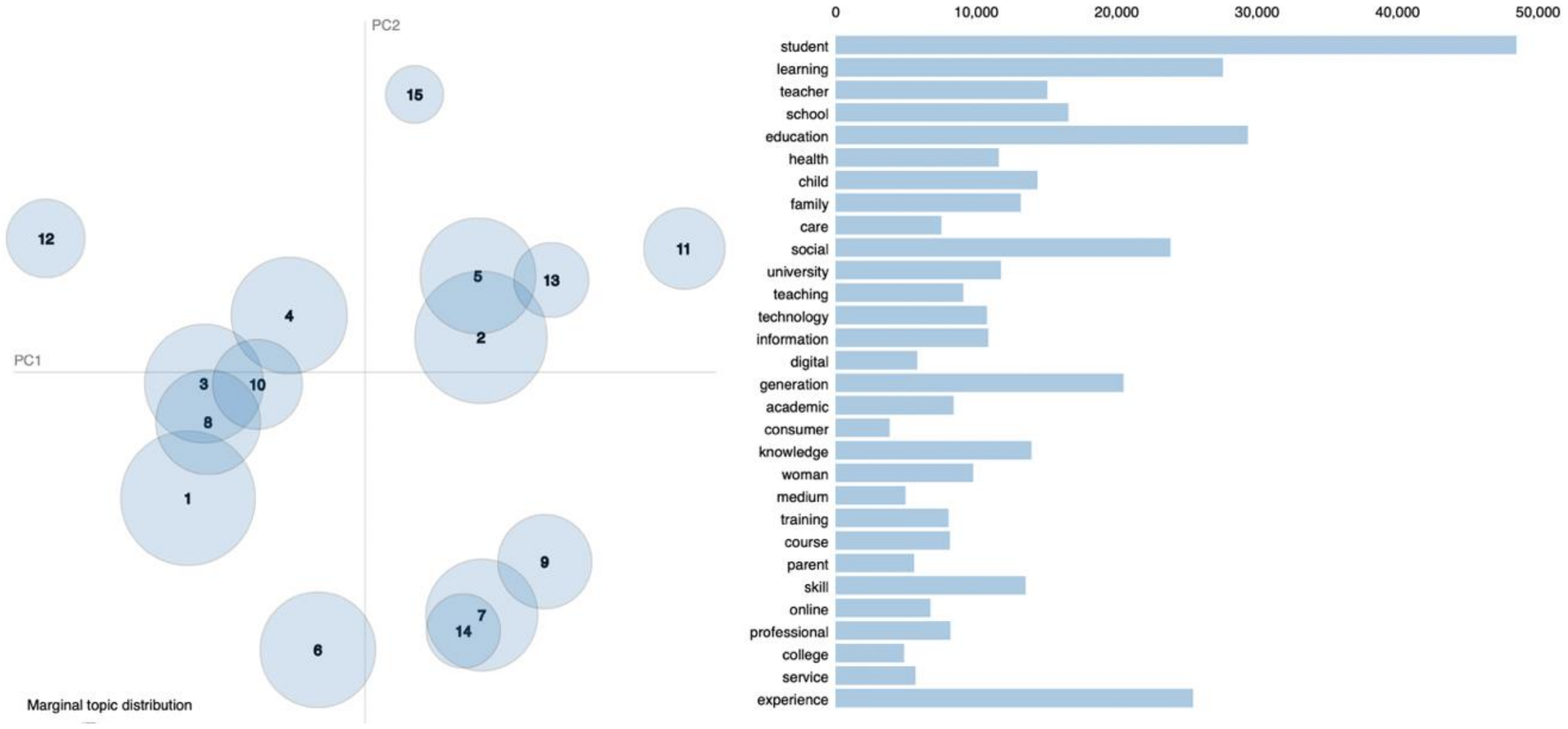

Figure 5 The Intertopic Distance Map of the topics and the most important words

In Figure 6, the histogram shows how many documents are in each topic. On the x-axis, the number of topics is displayed, while on the y-axis, the number of documents is displayed. As an example, Topic 12 has approximately 3500 documents.

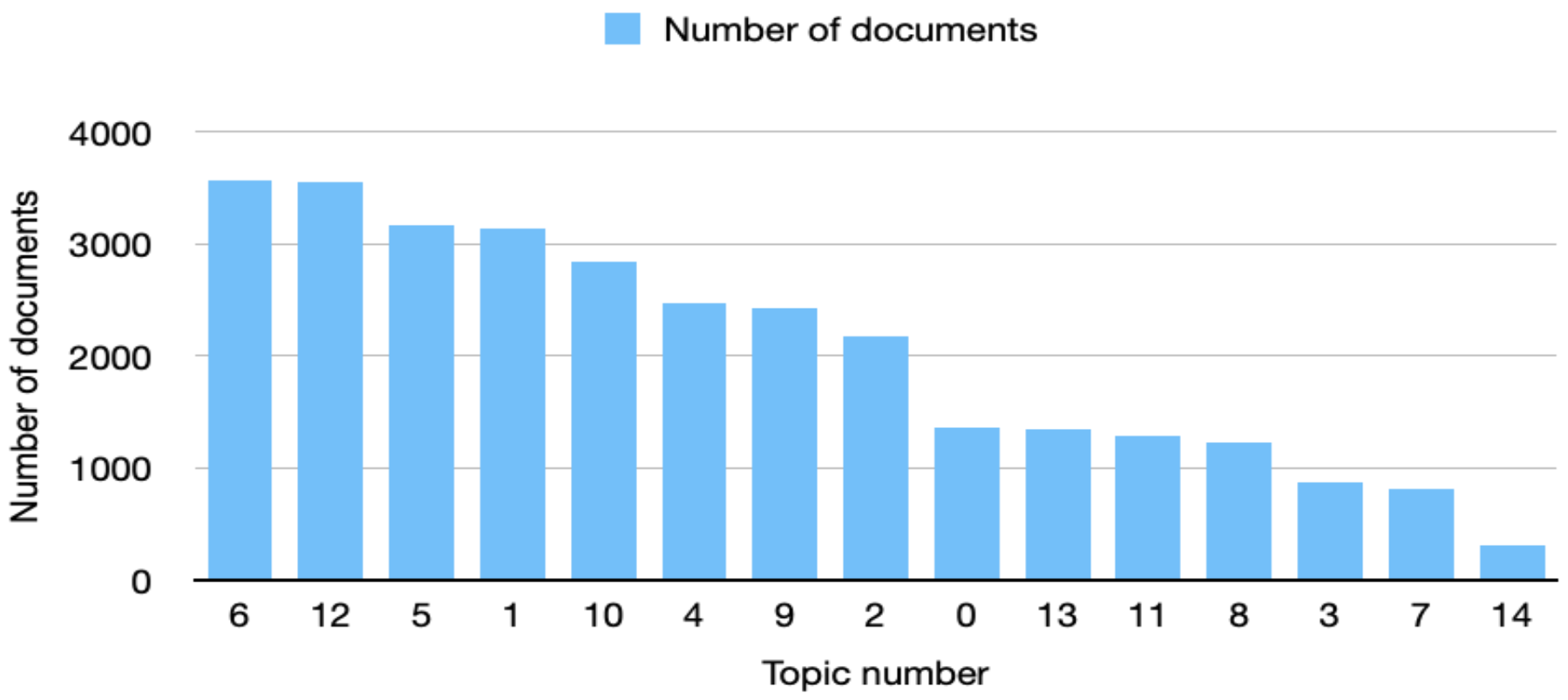

*Figure 6 Number of documents in each topic (Source: WoS)*

Figure 7 shows the document word count and the number of documents. Most documents in the dataset contain fewer than 500 words. A few contain between 500 and 750 words, and even fewer contain between 875 and 1000 words.

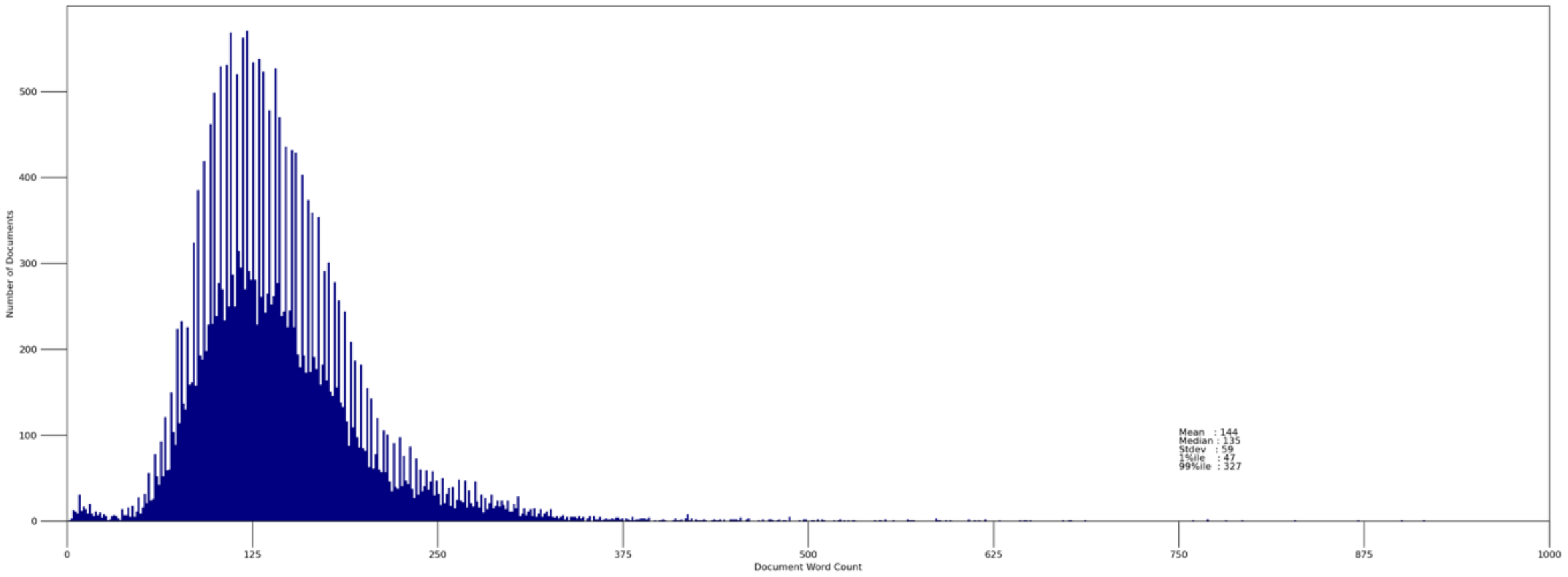

Figure 7 Histogram (Web of Science Article Abstracts)

Figure 8 shows word cloud for each topic. Based on frequency and relevance, the famous words and phrases in the articles are highlighted through Word Cloud.

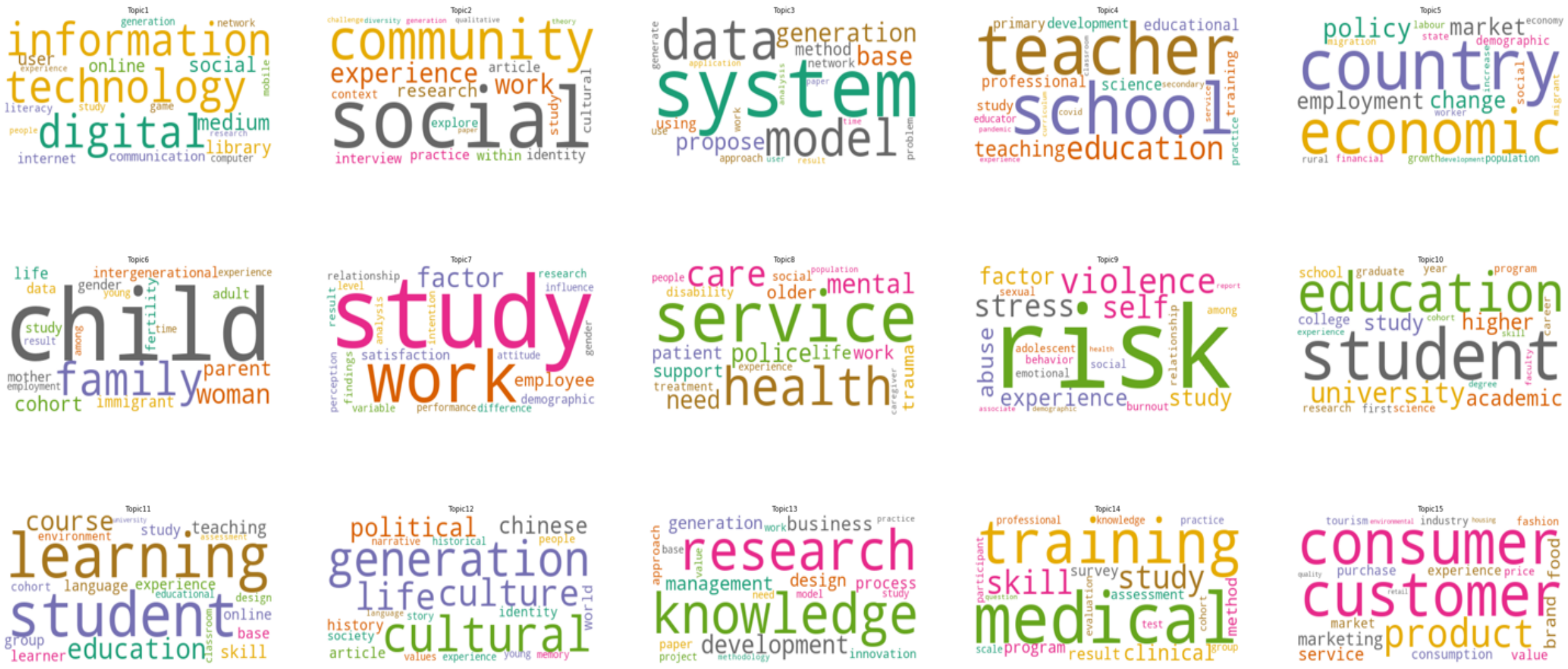

*Figure 8 Word cloud for each topic*

The ten most important keywords for each parameter are shown in Figure 9. Parameter keywords are indicated by vertical lines, and importance scores by horizontal lines. Colors indicate importance, with light blue being the most critical while dark blue being the least important.

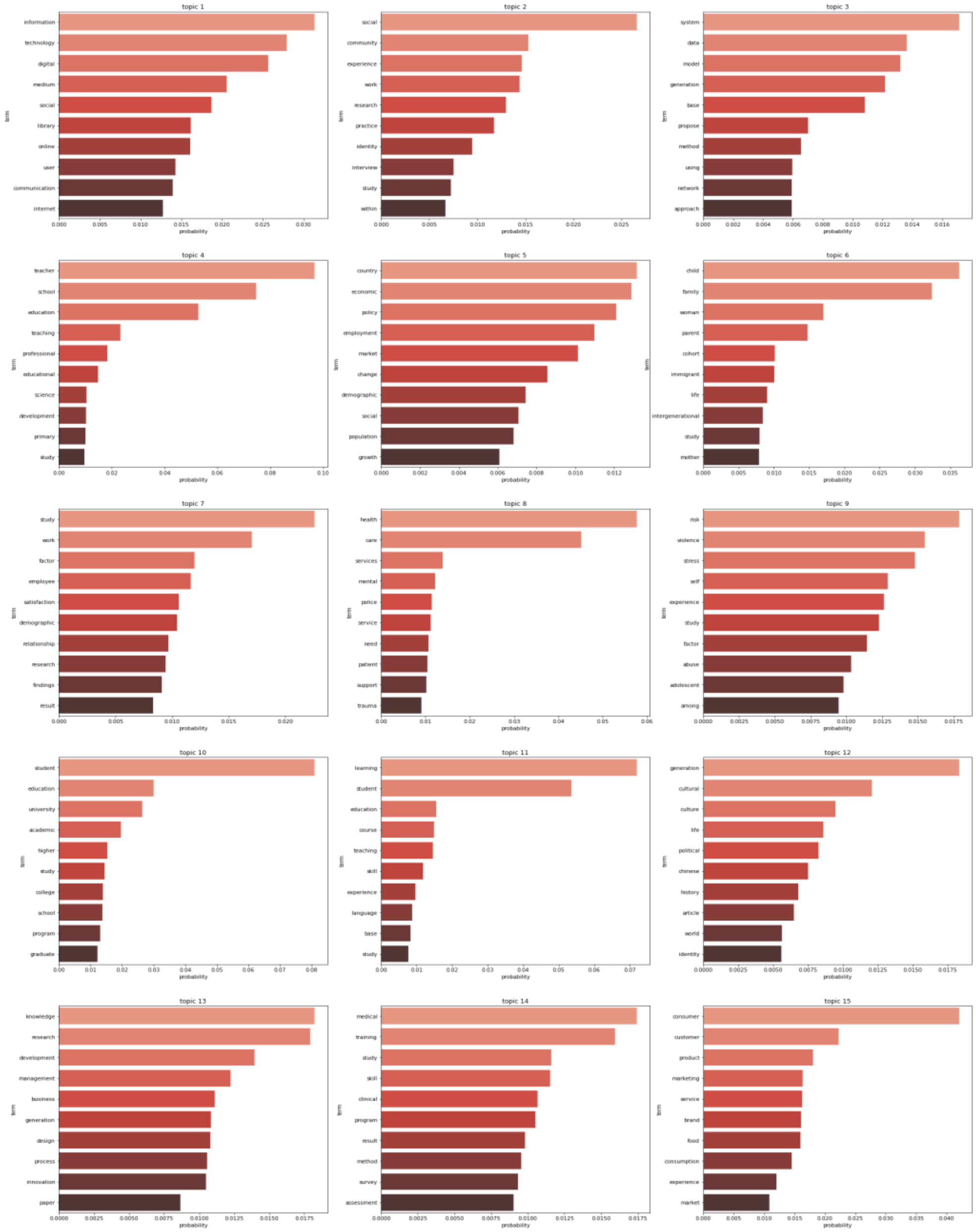

Figure 9 Term Score

### 4.3 Background on Generations Cohorts

A generation is a cohort of individuals grouped by age. They share that period's historical and social experiences, attitudes, and beliefs [38]. There are seven widely recognized generations: Greatest Generation, Silent Generation (Traditionalists), Baby Boomers, Generation X, Millennials (Generation Y), Generation Z, and Alpha Generation. Table 3 shows the different generations, their birth, and their ages. This section describes the different generations and their significant characteristics.

**Table 3** Different cohorts of generations

| Cohort | Year | Age in years (2022) |
|---|---|---|
| Alpha Generation | (2010-2024) | Less than 12 |
| Generation Z | (1995-2009) | 27 -13 |
| Millennials/Generation Y | (1980-1994) | 42 - 25 |
| Generation X | (1965-1979) | 57 - 43 |
| Baby Boomers | (1946-1964) | 76 - 56 |
| Silent Generation | (1925-1945) | 97 - 77 |
| Greatest Generation | (1901-1924) | 121 - 98 |

#### 4.3.1 Greatest Generation

The Greatest Generation [39] memorializes the Second World War generation by emphasizing on values they depict and instructing the audience to gain insight from this generation. This generation in the US came of age during the Great Depression before later fighting in the Second World War. It persevered through challenging circumstances attributed to war and economic stress. This nurtured a generation with the ability to cope with hardship and create a better world. The greatest generation’s members were born between the 1900s and 1920s, from 1901 to 1924. Their parents had the likelihood of being considered a constituent of the Lost Generation, and many of their children were within the Baby Boomer generation. Everly et al. [40] mentioned that the significant characteristics identified with this generation are personal responsibility, humility, excellent work ethic, frugality, commitment, integrity, and self-sacrifice.

### 4.3.2 Silent Generation

Lissitsa et al. [41] describe the Silent Generation as adults born in the mid or late 1920s, particularly from 1925, to 1945. The term “Silent Generation” was coined by Time Magazine in 1951 to describe the period’s emerging generation. Children growing up during this period were hardworking and kept quiet, with society believing that they needed to be seen but not heard [42]. They are cautious, unadventurous, withdrawn, and unimaginative [43]. Apart from conscientious and cautious behavior, the Silent Generation members are also considered thrifty, respectful to authority, and loyal to their careers, families, relationships, and religious beliefs. They are also described as being dependable and stable and having an affinity for stability. At the workplace, they are renowned for being team players, patriotic, and task oriented. Some of the essential values for the generation respect for authority, rules of conduct, and adherence to directions [44].

### 4.3.3 Baby Boomers

Baby Boomers are people born between 1943 and 1960 and were children of the early Silent Generation and Greatest Generation [45]. Boomers often highly regard education, and many depend on educational achievement to support their high urge for professional identity. Due to medical and scientific advances, Boomers are expected to live longer, thus denying the contemporary aging process [46] . Regarding career orientation, they are career focused and aspire to have a great career with the title, perks, and salary associated with it. Baby Boomers also tend to be highly competitive and seek value recognition and visibility. While they were not loyal to a single organization or firm compared to previous generations, they are not often “job-hoppers,” They often feel that changing jobs negatively affect their careers.

Furthermore, they value face-to-face interactions and prefer meetings within the workplace as the best communication mode. Baby Boomers tend to be exceptionally hardworking and motivated by prestige, perks, and position. They relish working for long hours and define themselves through professional achievements, with many feeling that Generation X and Y members should incur their dues due to their sacrifices. Many who have reached the working age have been found to possess a strong work ethic [47]. Studies have also demonstrated that Boomers are self-reliant, independent, and confident and feel that they have immense ability to change the world. Their courage makes them often challenge the status quo or question authority without fearing confrontation. Also, Boomers tend to be dedicated, achievement-oriented, and career-focused, and being competitive in the workplace enables them to enhance their skills. Wiedmer et al. [47] contends that many Boomers believe in the need to adhere to a hierarchical structure, thus making it difficult to cope with a flexible work environment.

#### 4.3.4 Generation X

Generation X consists of members born between 1961 and 1981. During this period, the US was characterized by severe economic recessions and the collapse of entire industries. Members of this generation grew up with limited adult supervision and thus learned the importance of work-life balance and independence. They are also highly educated, appreciate informality, are flexible and are technologically adept [48] . Employers highly value Generation X employees due to their position within the workforce. Whereas baby boomers continue to retire, and millennials are still attempting to ascertain their paths, Generation Xers can provide employers with a wealth of experience, stability, and knowledge. They are resourceful, individualistic, and self-sufficient because they have been used to taking care of themselves before they reach adulthood. They value responsibility and freedom and attempt to overcome challenges independently. Also, Generation Xers use their independence to advance their careers. Studies have also identified work-life balance as an aspect that Gen Xers excel in since they value a healthy balance between the time they spend at work and individual time and seek to pursue their aspirations. Because of their hate for personal possessions, Alferjany et al. [49] contend that members of this generation should be employed, rewarded, and handled differently compared to the earlier generations. In addition, they often have a robust entrepreneurial spirit and adapt to change well. Adaptability is a great asset for them, and if an employer initiates a new method or a client alters requirements, they do not experience any problem learning and succeeding. The first generation that grew up with digital technology, particularly computers, can adapt to technological programs quickly, unlike Baby Boomers, who mainly grew up with television. They find it easy to use several technological devices, including tablets, smartphones, and computers. While Generation X scores low in academic skills compared to Boomers, they have high scores in adult interaction skills, consumer awareness, and negotiating.

4.3.5 Millennials

After Generation X, studies are in consensus that the millennial generation was born between 1982 and 2003, a period of lowest ratio of children to parents in American history [49]. The rate of poverty among children who were less than six years also peaked in the course of this period [50]. Apart from nurturing and cherishing their children, Generation X and Boomer parents of millennials appeared to be obsessed with adequately preparing their children to be ready to face the future. Despite their pressure, millennials trust their parents and other authority figures. Contrary to their pessimistic Generation X or individualistic Baby Boomer parents, millennials tend to be team-oriented, optimistic, and confident. Like Generation X, millennials were born with computers and experienced fast Internet adoption. They are a highly connected networked generation and usually immersed in technology. Also, being career-oriented, they expect to advance rapidly and get great perks [51]. Millennials are used to being recognized for every achievement and being in the limelight. They appreciate feedback due to being evaluated, ranked, and graded throughout their lives. They value the involvement of their parents in their lives. Due to their multi-tasking skills, millennials aim to have parallel careers through continuous job-changing [52]. Community orientation also influences them to value mission-driven organizations. Millennials get motivated by assisting others and improving the world. They overly focus on learning and achieving throughout their lives, thus, are likely to value continuous learning opportunities.

Moreover, millennials like to participate in group activities and team sports and thus should be included in the organization's group activities. They crave attention, value praise, and feedback, and accept being guided by experienced mentors [50]. The seven core traits associated with the Millennials are high levels of optimism and trust, sheltered, unique, strong team instincts, pressure to excel, and conventional.

#### 4.3.6 Generation Z

Generation Z or post-millennials can be defined as people born from 1995 to 2012 [53]. They are the newest workplace participants and are primarily digital natives. Gen Z is also characterized by being achievement-oriented, seeking meaningful work, and are disengaged when there is no challenging work. Based on a systematic review conducted by Barhati et al. [53], extrinsic and intrinsic factors ascertain the career aspirations of Generation Z. In addition, based on predictions of past studies, the authors inferred that Generation Z has clear career development and expectations plans [54]. Members of Gen Z anticipate feedback on their work results. New technologies are natural for them. While they are willing to engage in foreign business trips, they are reluctant to relocate for work [54]. Also, they may not be loyal to employers but are ready to be employed by a single employer for prolonged periods, provided it is attractive.

#### 4.3.7 Alpha Generation

Gen Alpha is the generation following Generation Z, which now includes any children born between 2010-2024 [55], beginning from the year (2010) the iPad was born. They were born in an age where technology devices are becoming more intelligent and when everything is connected. As Alpha generation children grow, the new technologies will be a part of their lives, experiences, attitudes, and expectations about the world. They are at an increased risk of growing up to be selfish and expect immediate gratification. They are not in the labour force yet, but post-millennials impact the consumer market.

According to a recent report [56] from consulting firm McCrindle, by 2025, there will be over two billion members of Gen Alpha -- the most significant single generation ever.

### 4.4 Learning & Skills

The Learning & Skills macro-parameter includes four parameters: Management Science, Digital Learning Environments, Learning & Employment in Immigrant Families and Healthcare Sciences.

#### 4.4.1 Digital Learning Environments

The parameter Digital Learning Environments includes two clusters 3 and 11. Cluster 3 depicts the following keywords: system, data, model, generation, base, propose, method, using, network, approach, work, problem, use, generate, application, result, time, paper, analysis, and user. Cluster 11 detected keywords such as learning, student, education, course, teaching, skill, experience, language, base, study, group, learner, online, environment, cohort, design, educational, classroom, assessment, and university.

The parameter Digital Learning Environments includes Next-generation systems which are any system designed to work with new technologies, such as artificial intelligence [57] and the Internet of Things [58]. While these systems are often used to improve business processes and make them more efficient, they can also help analyze data, predict trends, automate tasks, and be used in learning environments. Digital Learning Environments requirements of Generation Z differ from those of previous generations. Teachers need to know how to use digital tools and spend more time on social media to connect with Gen Z students. Building collaborative E-learning environments [59] by using tools that make it easy for the instruction creator to design and construct complete collaborative settings. Also, allowing students to share their knowledge and experiences, this learning paradigm helped to increase the productivity of the online courses. Moreover, giving insights about Massive open online courses (MOOCs) [60]. Also, provides the importance of assessment tools for next generation students learning processes [61].

Also, Augmented reality learning to engage youth in urban environmental planning. Engaging the young generation in urban planning is a comprehensive strategy that acknowledges their crucial role in determining the future of urban life. Urban environmental concerns have been shown as untied relationships between communities and the young due to their passions for exploration, engagement with nature, and analysis.

### 4.4.2 Management Science

The parameter Methods for Generational & Employment Studies contains keywords such as social, community, experience, work, research, practice, identity, interview, study, within, article, cultural, explore, context, generation, theory, qualitative, challenge, paper, and diversity. Qualitative studies of the social, cultural, and personal aspects of society are, in essence, a study of the interaction between people. It examines the way groups, organizations and individuals interact with each other to form communities and networks and build relationships. This parameter explores human experiences, skills, and other aspects and how it has changed over time [62,63]. For instance, Qualitative interviews were conducted to explore the experiences of second-generation Asian Americans seeking employment [64].

### 4.4.3 Healthcare Sciences

The Healthcare Sciences parameter is characterized by medical, training, study, skill, clinical, program, result, method, survey, assessment, practice, cohort, evaluation, professional, participant, knowledge, test, scale, group, and question. The quality of healthcare is determined by all members of the healthcare team, not just the individual service providers; an approach to providing patients with the best possible quality of care through continuous assessment, training, evaluation, and improvement of the skills among healthcare practitioners. For instance, well-trained patient-centered physicians are known for their fundamental practices of communicating with patients [65]. To foster and sustain patient-centered attitudes, it is essential to guarantee job satisfaction and communication skills training.

On the other hand, Residency programs [66] can select medical students for interviews and employment using specific metrics. These include USMLE scores, GPA, and class rank/quartile. In addition, A validated search assessment instrument can effectively measure improvements in residents' search skills to demonstrate training efficacy and meet practice-based learning competency requirements of the Accreditation Council for Graduate Medical Education [67]. Also, the development of the nursing workforce is crucial to achieving the health-related Millennium Development Goals and would provide better medical care to China's rural population [68].

### 4.4.4 Learning & Employment in Immigrant Families

The parameter Learning & Employment in Immigrant Families is presented by keywords such as student, education, university, academic, higher, study, college, school, program, graduate, career, science, first, research, year, experience, skill, faculty, cohort, and degree. This parameter captures the academic issues of the first, second, and third generations of immigrants in the United States at different times. For instance, identifying the factors contributing to the retention of first-generation college students and facilitating this retention [69]. Also, the challenges faced by first-generation students that hinder the achievement of their academic goals [70]. Moreover, exhorting educational institutions to help students overcome barriers and become successful. This can be done, for example, by utilizing work-related employability training for a cohort of undergraduate business school students [71]. Also, demonstrating an incorporated curriculum in employability will likely improve students' employability decisions. In addition, analyzing the skills related across students of the first, second, and third generations in the United States [72]. Also, students who are first, second or third-generation students are often seen as less confident or less likely to do well in their education [73].

### *4.5* Employment Sectors

Employment Sectors macro-parameter includes two parameters: Education Sector, and Eldercare Sector.

#### 4.5.1 Education Sector

Our model detected the following keyword for the parameter Education Sector: teacher, school, education, teaching, professional, educational, science, development, primary, study, training, educator, practice, secondary, service, covid-19, experience, classroom, pandemic, curriculum. Preparing the next generation of teachers is a global priority [74]. Technology in schools and the excellent performance of so-called Technology Generation or Generation Z students are not adequate to promote digital competence. Technological and pedagogical instructor abilities are crucial [75,76]. For example, the potential benefits of using video games in the classroom include the ability to promote innovation, enhance critical thinking, and improve problem-solving. As their use increases, educators need to think critically about the impact a game-based learning environment can have on their students [77].

### 4.5.2 Eldercare Sector

The parameter Eldercare People includes health, care, mental, police, service, need, patient, support, trauma, older, life, work, disability, social, treatment, experience, people, population, and caregiver. In this parameter, we found the meaning of age has altered as Americans live longer, especially seniors and Baby Boomers. Thus, programs are created to increase students' civic involvement and professional growth, enable community debate on active aging, and prepare them for gerontology [78]. Moreover, baby boomers' health may be a significant factor in explaining their current workforce participation, with important implications for health expenditure and support in old age [79]. In addition, Gerontological social workers are needed to care for the aging population, according to social work authorities [80]. For this purpose, a gerontological social educator must help future, and current practitioners understand how to help their patients live a fulfilling old age. Observing students and practitioners demonstrate competence in carrying out taught skills is essential to ensuring this outcome [81].

## 4.6 Consumer Industries

Consumer Industries macro-parameter includes only one parameter: Marketing & Consumer Behavior.

### 4.6.1 Marketing & Consumer Behavior

The keywords detected by our LDA model for this parameter are consumer, customer, product, marketing, service, brand, food, consumption, experience, market, purchase, value, industry, fashion, tourism, price, environmental, retail, housing, and quality. The Marketing & Consumer behavior areas of study focus on how consumers make purchasing decisions. This includes everything from demographics and different generations to attitudes to social media usage [82–84]. For instance, Indian consumers' definition of retail service quality concerning small retailers. However, it did not consider demographic variables like income, education, and occupation [85]. On the other hand, comparing consumers from different generations within a target segment – bank service customers [86]. Also, differences between tourist satisfaction levels according to their demographic characteristics: gender, age, educational level, and employment status [87]. Moreover, the issue of the growing number of elderly customers, baby boomers who retire, become more demanding of high-quality goods and services. However, they also have more difficulty adjusting to the store environment and shopping [88] .

## 4.7 Learning & Employment Issues

Employment Issues macro-parameter includes three parameters: Multi-generational Workforce, Employee Satisfaction, and Mental Health.

### 4.7.1 Multi-Generational Workforce

The parameter Multi-Generational Workforce describes the workplace environments across different generations. It is represented by keywords including knowledge, research, development, management, business, generation, design, process, innovation, paper, approach, project, model, base, work, study, need, value, practice, and methodology. Multi-generational workforce may be vital to creating an adequate knowledge working environment and contributing to the goal of organizations being more productive and competitive [89]. For example, Highlighting the values and work ethic of Generation Z while briefly describing the characteristics of the Baby Boomers, Generation X, Generation Y, and Generation Z[90]. The labour market is examined by studying the behavior of Generation X and Generation Y (Millennials) [91]. In the foreseeable future, it is anticipated that these two generations will significantly shape the labour market. On the other hand, the primary abilities and knowledge of a new young generation of Library Information Services (LIS) workers has been discussed [92]. Also, an overview of the digital world and its effects on LIS workers has been provided. Moreover, research on age-friendly workplaces focuses on the physical work environment, positive organizational attitudes, and the link between employee experiences and performance [93].

### 4.7.2 Employee Satisfaction

Keywords that represent the Employee Satisfaction parameter include example, study, work, factor, employee, satisfaction, demographic, relationship, research, findings, result, performance, difference, influence, gender, attitude, variable, level, intention, perception, and analysis. Research showed different characteristics associated with each generational cohort in the workplace. Generation Y and Z are far more likely to be competent in technical fields. There is no doubt that generation X is technically behind gen Y. To increase job satisfaction, we must understand how challenges, issues, and characteristics work [94]. There was a difference between Baby Boomers and Generation X workers in job satisfaction, organizational commitment, and willingness to leave the workplace. According to the results, boomers are more likely to be satisfied at work and are less likely to quit than Xers [95]. Moreover, statistics showed that job satisfaction positively correlates with age [96].

### 4.7.3 Mental Health

The parameter Mental Health is characterized by keywords such as risk, violence, stress, self, experience, study, factor, abuse, adolescent, among, social, sexual, emotional, behavior, relationship, burnout, demographic, health, report, and associate. Many people and especially young ones, feel overshadowed and the feeling of being constantly watched and judged. They also lack experience with bullying, family dysfunction, or emotional abuse, making them vulnerable to depression [97]. All these things can cause strain on their self-esteem, making it more likely for them to develop depression symptoms. In addition, they have new responsibilities like making money, finding jobs, and school; increased stress from all of that; and often less support from family members who are moving on with their lives or are not around at all [98]. This can lead to feelings of isolation and helplessness, a particular concern in older people [99]. In addition, workplace bullying is a form of workplace mobbing, the negative behavior exhibited by coworkers that can cause severe psychological distress [100]. In many countries, it is illegal for an employee to bully others at work - including via electronic means such as social networking sites.

On the other hand, smartphone addiction has been widely studied in recent years. Smartphones could affect various demographic, personality-linked psychological, and emotional variables [101]. Thus, the study to understand the elements that influence smartphone addiction across generations may lead to more effective education, productivity in the workplace, and awareness efforts on technology's impact on mental health. Also, smartphone users from Generations X, Y, and Z workers answered an 80-item questionnaire [102]. Anxiety over smartphone use was measured by characteristics such as social pressure and emotional gain. The study's key finding is that Generation Y has much more addictive behavior than the other two generations. The most significant predictors for all three generations were social environment pressure and emotional gain. Interestingly, emotional gain from smartphone use was more significant for generation Z than for older generations. Like neuroticism and everyday usage time, WhatsApp usage was a significant predictor for Generation Z.

## 4.8 Generations-specific Issues

Generations-specific Issues macro-parameter has three parameters: Technology Usage, Culture & Identity, and Immigrant Families & Intergenerational Life.

### 4.8.1 Technology Usage

is represented by keywords detected by our model such as information, technology, digital, medium, social, library, online, user, communication, internet, literacy, game, generation, mobile, computer, study, network, experience, people, research. The parameter talks about the digital age [103–106], which has changed how we communicate, learn, work, and consume information. It also discussed the challenges for people who are not digitally savvy, such as older people (baby boomers). For instance, Mobile phone penetration has changed the traditional dynamics of older people leading. Now they depend upon young people such as Gen Y and Z to meditate to help them use mobile phones [107]. Positive interactions increase the provision of care by young people to older people. Moreover, it can utilize smartphones and other technologies to enhance the learning outcomes of Gen Z students who are already proficient in using such technologies [108]. Also in book preferences, the preference for print and electronic books for reading purposes has many factors like gender, age, education level, income, ethnicity, etc. For example, Baby boomers and Gen x are more likely to prefer paper books over electronic ones [109]. Also, examining smartphone use categories, process and social, for connections with depression and anxiety and problematic smartphone use [110]. In addition, examining the relationship between process and social smartphone usage, emotional intelligence, social stress, self-regulation, gender, and age. For instance, age influences process, social usage, and stress. Older people are less likely to acquire smartphone addiction [111].

### 4.8.2 Culture & Identity

The parameter Culture & Identity includes many keywords such as generation, cultural, culture, life, political, Chinese, history, article, world, identity, society, people, memory, young, values, experience, story, narrative, historical, and language. There are many different cultures in the world, each with its unique history and heritage. In order to understand and explain our current identity in our culture, the work of previous generations can be studied [112], [113]. For instance, the fundamental differences between Generation X and Baby Boomers regarding cultural traits and work ethics [114]. On the other hand, the current generation enters a discourse with history and decides what needs to be kept or added and what needs to be discarded. For example, Chinese co-ethnics from the first world who work in the People's Republic of China may uniquely use their Western culture and schooling and are presumed to understand Chinese culture [115].

### 4.8.3 Immigrant Families & Intergenerational Life

We combine two clusters, numbers 5 and 6, into a single parameter. The Immigrant Families & Intergenerational Life parameter is represented by keywords, child, family, woman, parent, cohort, immigrant, life, intergenerational, study, mother, adult, fertility, gender, data, employment, experience, among, young, time, and result. The parameter includes research about the issues faced by immigrant families who come to live in the United States or any other country from their original one [116–118]. For example, A comparative study is conducted between women and their descendants, looking at whether family formation differentially affects their labour market position. Regarding employment levels, migrant women of first and second generations have higher unemployment rates than native women, regardless of background characteristics, suggesting differential employment opportunities and family policies among these groups [119]. Also, the intergenerational mobility pathways across migrant and non-migrant families in the education and occupational domains are examined [120]. On the other hand, the solidarity of a family and a healthy intergenerational relationship between a grandparent and grandchild is important [121].

**5. Parameter Discovery for Multi-Generational Labour Markets (LinkedIn)**

Twitter and Facebook have been the subject of intensive research over the past few years, and datasets are readily available. LinkedIn is the largest network of professionals and hence it is an exciting source of data to explore. However, LinkedIn does not have a dataset gathering information (posts) and hence it is challenging to develop datasets from it. LinkedIn was our data source to conduct a study focusing primarily on labour markets involving professionals. A web scraping and social mining technique was used to obtain the dataset. This section presents the parameters detected using our LDA model based on the LinkedIn dataset. We briefly describe in this section the parameters and macro-parameters discovered from the LinkedIn data and use them to develop the multi-perspective taxonomy of multi-generational labour markets. A detailed description of the parameters can be found in [7].

We detected 15 parameters. Two of them were merged into one. One of them was discarded since it was in a different language. So, we are left with 13 parameters. These parameters are divided into five macro-parameters. The parameters are categorized into five macro-parameters: Learning & Skills, Business & Employment Sectors, Consumer Industries, and Learning & Employment Issues and Generations-specific Issues macro-parameter has one parameter, which is Crimes & Racism. Learning & Skills macro-parameter has two parameters: Learning & Skills and Leadership. Business & Employment Sectors macro-parameter has four parameters: Remote Work, Recruitment, Entrepreneurship, and Family Business. Consumer Industries have six parameters: Brand Marketing, Retirement, Energy Sectors, Entertainment, Celebrations, and Mental Health. A taxonomy of the multi-generational workforce is shown in Figure 10.

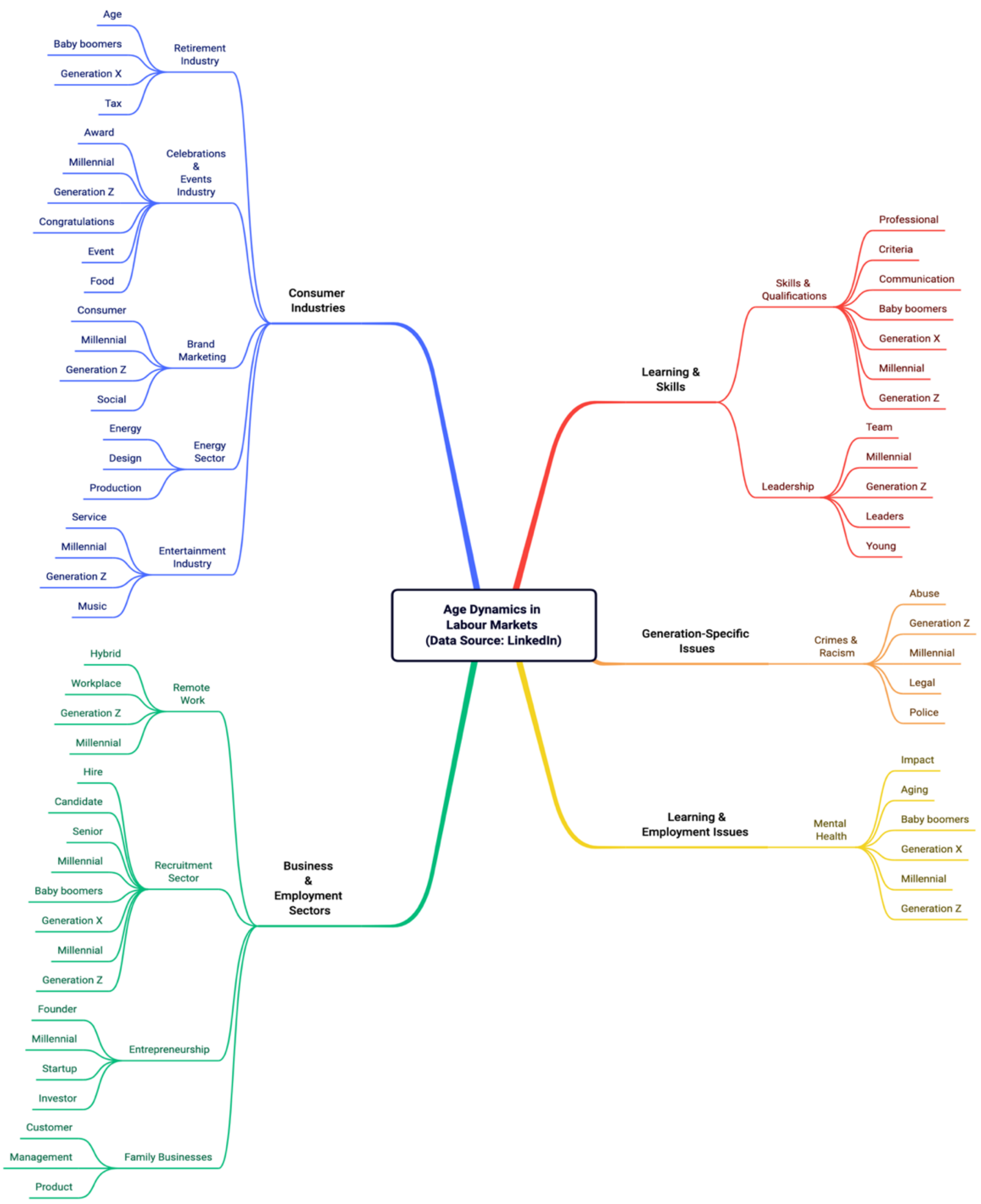

Figure 10 LinkedIn perspective taxonomy of Multi-generational Labour Market

The parameters addressed numerous issues in the labour market. The millennial generation, for instance, should understand the challenges associated with working from home and how to overcome them despite the benefits remote work offers. A significant challenge of working from home is the increased risk of burnout. Office environments give employees time to decompress and recharge after a long day at work. Employees who work from home are not afforded this luxury, so they are often under more pressure to be productive constantly. As a result, burnout can occur, and high-stress levels can negatively impact health. To overcome that, it is essential to take breaks throughout the day. This way, employees will be energized and ready to tackle another productive workday. Additionally, mental health needs to be taken care of because it can affect physical health and relationships. To feel comfortable speaking with friends and family members about their feelings, seniors and adults should feel free to express themselves.

Our results also touched on hiring and retaining employees, which is becoming more challenging for organizations. In addition to managing their workforce effectively, organizations have difficulty hiring and retaining the right employees. Depending on the organization, multi-generational employees may leave due to a lack of advancement opportunities. Additionally, creating a culture that appeals to all generations can be challenging. As a result, organizations may need to adjust their policies to be more inclusive of different generations. In addition, flexible work schedules, relevant training programs, and open communication between generations can enhance employee satisfaction.

In addition, we found that the number of retirees in the United States will outnumber children by 2023. The workforce will be significantly affected by this change. There will be a decreased labour pool, resulting in higher employee demands and increased costs of hiring, training, and finding new employees. Offering part-time schedules and flexible scheduling options to retirees is one solution.

In addition, we found that unemployed Generation Z and Y youth are five times more likely to have criminal records. Furthermore, if racial issues occur at work, they can severely affect both the victim and the workplace. There is a possibility that victims will feel uncomfortable or unwelcome in their jobs, and they may lose respect for colleagues and co-workers. Employees can also distrust each other in a racist environment, which hinders teamwork. Employees with racist attitudes are also more likely to be unproductive and leave the company more often.

Despite challenges in the 21st-century economy, businesses are in an excellent position to remain competitive. The reason for this is that entrepreneurs can be of any age. An entrepreneur can start a small business at any age, or an employee at an established company can start a business. Also, they can consider investing in startups or engaging in other entrepreneurial activities. Although entrepreneurs will have many challenges ahead, they must be prepared for them. Additionally, from the perspective of the brand marketing of the business, it was found that modern digital technology and media are required to meet the needs of youth customers (Gen Y and Gen Z).

## 6. Discussion

In this paper, we proposed a data-driven artificial intelligence (AI) based approach to automatically discover parameters for multi-generational labour markets using academic literature and social media analysis. Specifically, we discover parameters from Web of Science and LinkedIn posts using the Latent Dirichlet Allocation algorithm (LDA). We developed a software tool from scratch for this work that implements a complete machine learning pipeline using two datasets.

We discovered 15 parameters within the Web of Science dataset and categorized them into five macro-parameters: Learning & Skills, Employment Sectors, Consumer Industries, Learning & Employment Issues, and Generations-specific Issues. Figure 11 shows the word cloud of keywords discovered from Web of Science articles, with the size of each keyword denoting its frequency, a measure of its importance. According to the figure, the Web of Science is primarily concerned with work, education, experience, research, and generation.

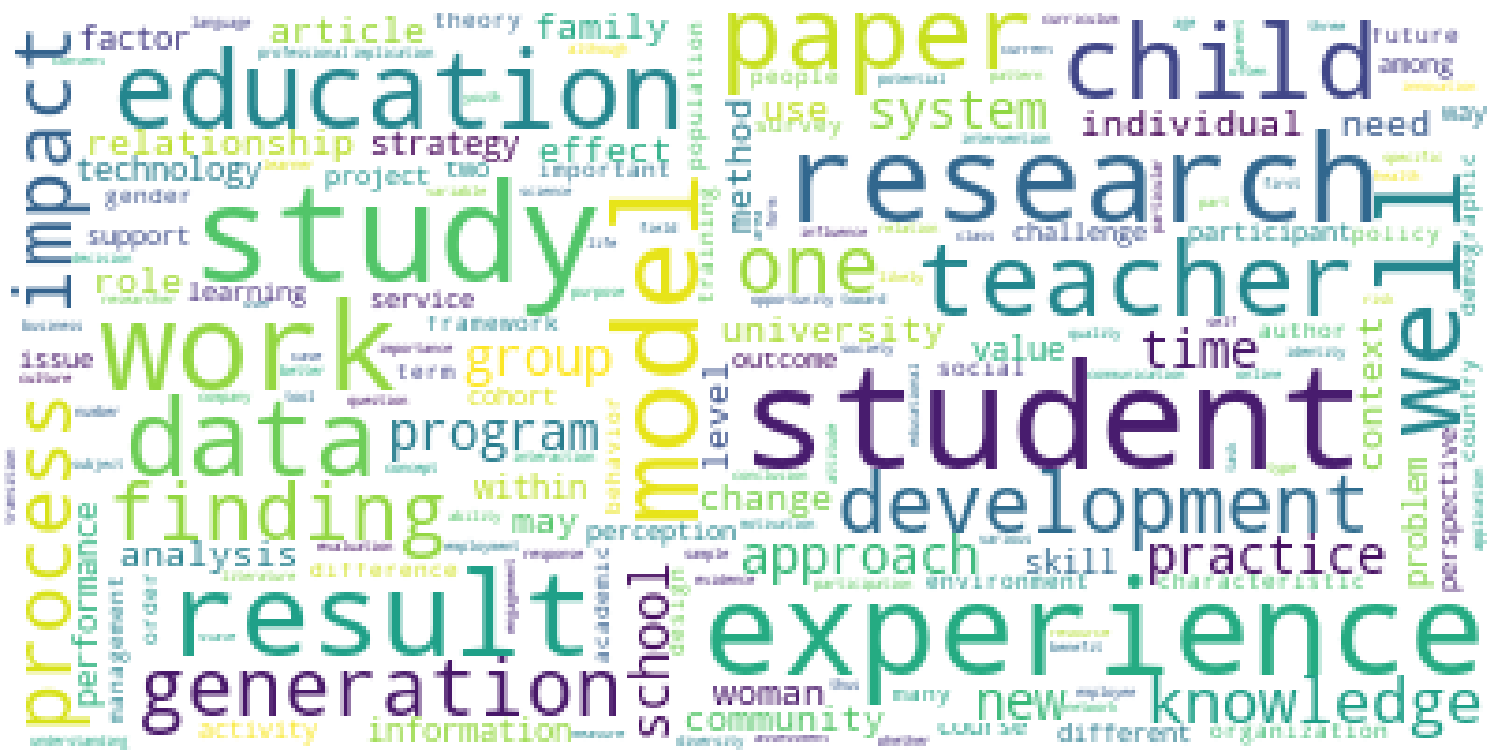

Figure 11 A Word Cloud generated from Web of Science

Using the LinkedIn data, a total of 13 parameters were discovered and categorized into five macro-parameters. The macro-parameters are the same as for the Web of Science, however, there are differences in their constituent parameters. Figure 12 shows the word cloud of keywords discovered from LinkedIn posts, with the size of each keyword denoting its frequency, a measure of its importance. According to the figure, LinkedIn is primarily concerned with work, education, research, and generation. These foci though are somewhat similar to the Web of Science the overall mix of foci is different.

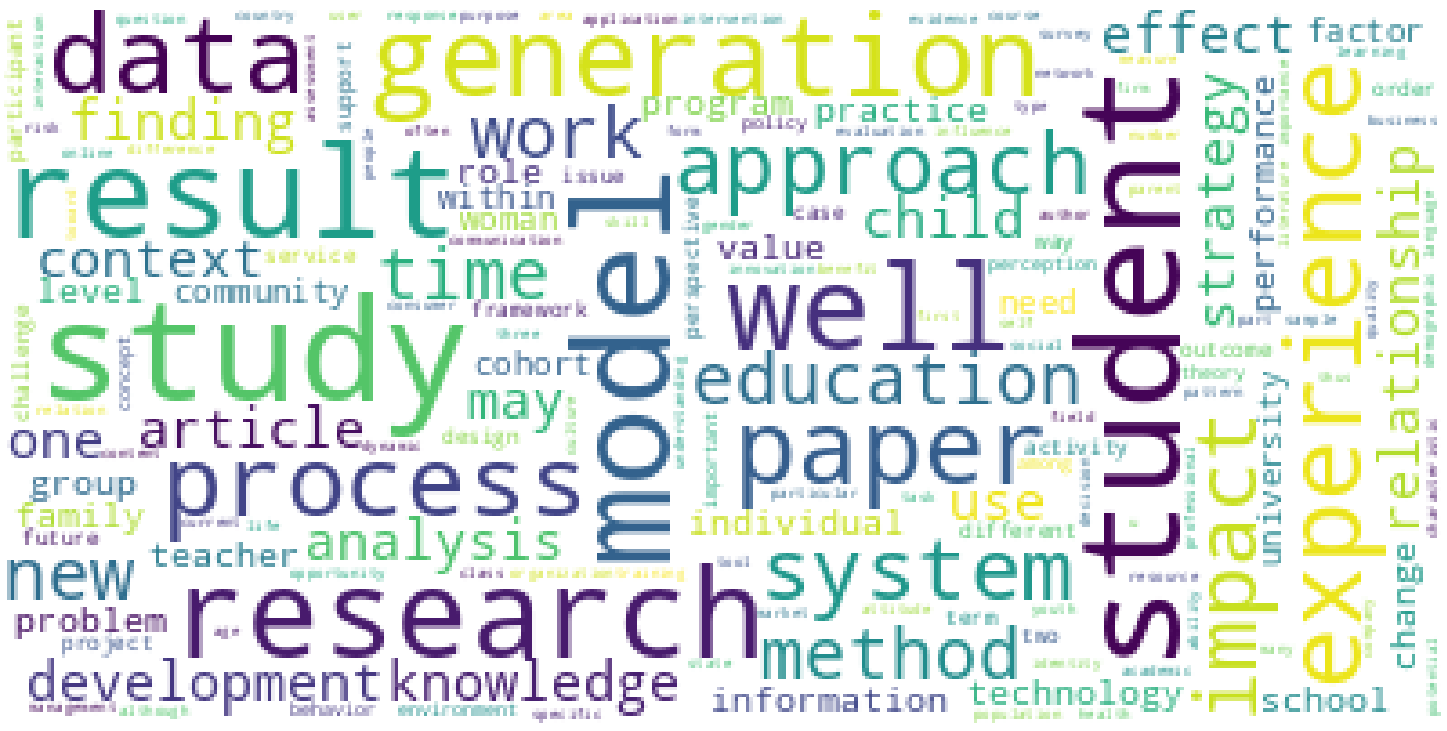

Figure 12 A Word Cloud generated from LinkedIn posts

Figure 13 shows a multi-perspective (academic and professionals) view of the generation characteristics in labour markets discovered by our tool. We depicted the importance of hiring and retaining employees. In order to hire and have the right employees, organizations may need help managing their workforces. Several generations may leave the workplace due to a lack of advancement opportunities. Creating a culture that appeals to all generations is also difficult. This issue may require organizations to make their policies more inclusive of different generations. Among the things that can be done are offering relevant training, offering flexible work schedules, and establishing an open line of communication between generations.

Also, we found that baby boomers in their 50s and 60s are retiring, and more retirees will live in the US in 2023 than children. As a result, the workforce will be significantly affected. With a shrinking labor pool, hiring and training new employees will become more expensive as there are fewer qualified applicants available. A possible solution is to hire more retirees and offer them part-time work.

Moreover, unemployed American youth (Gen Y and Gen Z) have five times more criminal records than employed youth. Moreover, racism at work can harm both the victim and the workplace. As a result of abuse, victims may feel unwelcome at work and lose respect for their coworkers. In addition to discouraging teamwork, racist workplaces can also foster distrust. It is also possible for racist attitudes to decrease employee productivity and turnover.

On the other hand, Today's businesses are well-positioned for success in the 21st century. It is possible to start a small business or to work for an established company at any age since entrepreneurs can start businesses at any age. Other entrepreneurial activities include investing in startups or participating in other entrepreneurial ventures.

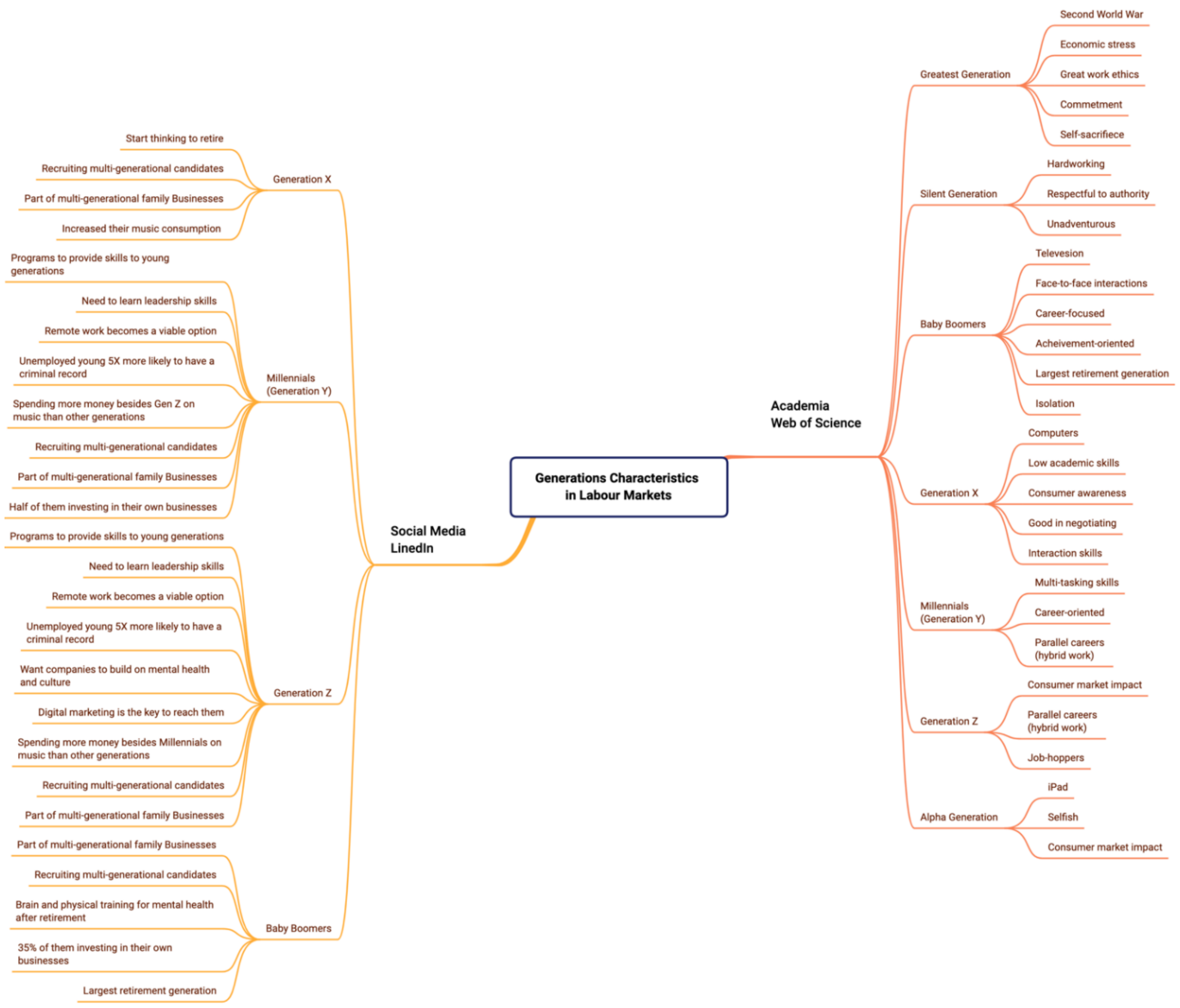

Figure 13 A Multi-Perspective (Academic and Professionals) View of the Generation Characteristics in Labour Markets

## 6.1 Multi-Generation Workforce Skills and Benefits and Challenges

We discuss here the skills and benefits of multi-generational workforce derived from the discovery and analysis of parameters conducted in this study (consider also the multi-perspective view depicted in Figure 1). In the past decade, technology has profoundly impacted the workplace. It has transformed how we work and interact with one another, how we manage our time and how we complete tasks [122]. This transformation is particularly evident when examining how technology has changed how employees perform their jobs. These changes are especially apparent when examining how multi-generational workplaces have evolved and continue to devolve into a new normal [123]. In today's multi-generational workforce, different attitudes and experiences among baby boomers, Gen Xers, Millennials (Generation Y), and Gen Z present greater variety in adopting and using technology [124]. With more generations working together, we see wider differences in how various age groups view technology, processes, and workplace productivity. Younger generations may share technology skills, while older generations impart sectoral knowledge and general wisdom in the workplace [125]. The millennial generation is now entering the workplace in droves [122,126], bringing their distinct Millennial values with them as they take on leadership roles and manage other employees who may be from an older generation or younger generation than themselves. As it stands today, millennials are the largest generation currently in the workforce [127].

On the other hand, the shift in technology and paradigm among employers also reflects employee trends. The email and phone poll responses seemed unanimous; these trends are not going away anytime soon because technology is being used to help solve problems in the workplace and develop growth opportunities within companies from all sectors. Employees from different generations can now work collaboratively from remote locations and meet virtually without ever having met face to face throughout history, with systems like video conferencing furthering their capability to interact virtually for remote meetings.

Currently, employees no longer retire at 65 or remain with the same organization for decades. This suggests that the days when employees used to work their way up the ladder are long gone. With such a diverse talent pool, businesses can garner a competitive edge through the implementation of a multi-generational workforce. With the workforce becoming increasingly diverse, it is crucial to comprehend the advantages of an inter-generational workplace. Conceptually, an inter-generational workforce is one that has employees from various age groups, including Baby Boomers, Generation X, Millennials, and Generation Z. A multi-generational workforce is defined as employees made of individuals emanating from diverse generations. This section focuses on the definition, benefits, and challenges of a multi-generational workforce.

A multi-generational workforce provides an organization with a plethora of benefits. The first advantage identified in the literature is that it can result in more effective communication since employees of diverse age groups have different ways of communication. For example, Generation Z and Millennials may prefer communicating through messages or social media whereas the Generation X and Boomers may prefer face-to-face communication or emails [126]. Organizations can tap into the several styles of communication that the different age groups use to disseminate messages more effectively. Because of the diversity of the workforce, organizations can grab this as an opportunity for employees to gain insight from each other and enhance communication across the various generations. A multi-generational workforce can result in better communication due to the various communication styles of the different generations. Understanding and respecting such differences can be critical for bridging the gap and enhancing understanding between employees [45]. A cohesive working environment that such a workforce has can be beneficial in improving the overall productivity of the organization.

The second benefit of an intergenerational workforce is that it can offer various experiences and skills. Employees of various ages usually have varying experiences and skills to offer. For instance, Millennials and Gen Z may have more familiarity with new technologies, whereas Gen X may be experts in traditional methods [128]. Mixing the various age groups within the organization can enable them to benefit from various experiences and skills. The other advantage is better creativity and innovation for meeting the dynamic marketplace needs. Different age group employees bring various experiences and perspectives on things such as product development, work processes or customer service, thus resulting in innovative and exciting solutions to issues when companies tap on them [129]. An intergenerational workforce can also assist in creating an innovation culture since employees feel free to share their ideas. In this regard, businesses which adopt a multi-generational workforce have more likelihood of being successful in the current dynamic marketplace since they have better innovation and creativity that guarantees them a competitive advantage.

Better adaptability and flexibility have also been cited as a critical benefit since employees of various age groups often have varying skill sets. For instance, Millennials and Gen Z are more tech-savvy and embrace new technologies quickly [130], whereas older generations such as Generation X can have more institutional knowledge and experience. Having a multi-generational workforce can ensure that the organization becomes more adaptable to change and flexible. Moreover, a multi-generational workforce can result in better problem-solving. With a combination of backgrounds and ages, employees can introduce various perspectives for solving workplace problems, thus resulting in more effective and innovative solutions. The more well-rounded team that emanates from a multi-generational workforce can enable an organization to tap into a wider pool of knowledge and skills. It can make the team to become more versatile and meet the workplace challenges [131]. Younger workers can gain insights from experiences of older workers, thus help in creating a more supportive and collaborative work environment. By creating a dynamic workplace, a multi-generational workforce increases the likelihood for more exchange of ideas and vibrant work culture. Increased profitability and productivity are the other benefits of a multi-generational workforce. Employees of different backgrounds and ages can introduce various viewpoints on workplace issues [132]. The more potential for exchange of ideas among the different backgrounds and ages can result in better profitability and productivity within the workplace.

### 6.2 Multi-Generation Workforce Challenges

We discuss here the challenges that a multi-generational workforce poses. These are derived from the discovery and analysis of parameters conducted in this study.

Notwithstanding the numerous benefits accrued from a multi-generational workforce, it also poses certain challenges. One of the major challenges is the management of various expectations on work-life balance. For instance, Generation Z value work-life balance more than Generation X. In this regard, they are likelier to leverage on flexible work arrangements such as compressed workweeks or telecommuting [133]. As a generation that directly encountered the great recession, Generation Z are more concerned about job security, perks, and salary, but at the same time tend to be vocal regarding work-life balance as well as workplace flexibility. Notably, work-life balance has been found to be quite pertinent not only for Generation Z but all other employees [133]. In fact, this generation holds the assumption that it is the mandate of the organization to offer flexibility since it only increases efficiency and productivity [134], and that such a work arrangement needs to be open. Balancing such interests with the other demands of other generations regarding work-life balance may be quite challenging for an organization.

Communication is an important tenet for the workforce. However, different generations have different communication styles. Whereas Gen Xers often prefer emails and calls, Generation Y often prefers to send instant messages. The use of informal language, abbreviations, and colloquialisms further contributes to communication breakdown, thus making communication a major challenge. To address this, organizations should encourage collaboration among the different generations to ensure they learn from each other [135]. Changing the collective mindset and influencing different generations to view themselves as partners can ensure everyone benefits from new communication forms and ideas. Motivating a multi-generational workforce is also quite daunting. Motivating employees usually implies the creation of a corporate culture that supports all persons' goals and ideals. It can include flexible working strategies and perks that enable various generations to pursue diverse aims within their careers and work in several ways to actualize them. To address this, it is crucial for multi-generational workplaces to treat each employee as an individual. Instead of motivating all persons with similar benefits, an organization should personalize its motivational approach [135].

Every generation has certain stereotypes which may be harmful to the workplace. For example, older workers like Gen X often believes that Millennials and Gen Z are entitled and tech-obsessed, whereas the younger employees such as Gen Z believe that Baby boomers and Gen Z are stubborn and old-fashioned [136]. This illustrates that while various generations embrace different styles and preferences of work, stereotypes can negatively affect the work environment. To tackle this, it is crucial to concentrate on valuing employees for their individual strengths [135]. Employers should not assume that certain individuals in the team require special treatment and help. They should not concentrate on weaknesses of other workforce members. Rather, they need to know each employee individually, and concentrate on tapping on their strengths.

Furthermore, it is difficult to balance the weaknesses and strengths of a multi-generational workforce. Every generation has its characteristics that it delivers to the workforce. Such differences need to be embraced to enable organizations leverage on their teams. However, managers at times view the gaps that exist between members of teams as negatives. Despite this, developing a team that brims with diverse perspectives and insights can be beneficial to an organization [124]. To tackle this, cross-generational mentoring can play a pivotal role, and this can be done through the development of a reciprocal mentoring program. Such a program can ensure, for example, that Millennials teach Gen X how to utilize social media along other forms of technology, whereas Baby Boomers can guide Gen Z into communication and interpersonal skills while sharing knowledge regarding the way the business functions. The organization should allow team members to gain insights from each other and depend on other organizational members when they need assistance to balance their weaknesses and strengths. In addition, the organization should fine-tune onboarding programs to meet the needs of the various generations, especially Generation Z [134]. A well-designed program can enable new hires to minimize uncertainty and anxiety and provide knowledge and clarity to their role. Effective onboarding can also lead to better job satisfaction, loyalty, and performance. Mentoring can be instrumental in ensuring the younger employees coordinate with various organizational processes.

## 7. Conclusions

Capitalist approaches have had negative impacts on social, economic, and environmental sustainability. Economic issues such as inflation and energy costs have been exacerbated by global events and financial crises have revealed weaknesses in modern economies. The current trend of people quitting their jobs in large numbers, known as the Great Attrition, and the presence of multiple generations in the workforce also pose challenges. Transformative approaches are necessary to address these issues and protect society, the economy, and the environment.

This study uses big data and machine learning to identify multi-perspective parameters for multi-generational labour markets. The parameters were discovered using 35,000 academic articles and 57,000 LinkedIn posts, and were organized into 5 macro-parameters: learning and skills, employment sectors, consumer industries, learning and employment issues, and generation-specific issues. The study also includes a knowledge structure and literature review of multi-generational labour markets based on over 100 research articles. A machine learning software tool was developed for data-driven parameter discovery and various quantitative and visualization methods were applied to explore the topic.

This paper presents an approach for obtaining comprehensive, objective, and multi-perspective information on a subject using machine learning and deep learning, and provides tools and resources for accessing information from various datasets. The research in this paper contributes to our understanding of age dynamics in labour markets and may be used to raise awareness and drive future research on the topic using advanced technologies. The findings and knowledge gained from this work can be used to inform decisions and guide labour economics research, and the work is expected to enhance the theory and practice of AI-based methods for information and parameter discovery, extend the use of LinkedIn and scientific literature media for information discovery, and promote novel approaches to labour economics and labour markets. Ultimately, this work aims to contribute to the development of sustainable societies and economies.

Artificial intelligence is enabling autonomous functionality in various systems, including self-driving cars and robots, and will be extended to larger systems such as industrial sectors and governance. By identifying and defining the key characteristics and variables that will guide the design and operations, designers and managers can ensure that they are creating a product, system, or process that meets the desired specifications and requirements. The same concept is applicable, though much more complex and grander, to the design and governance of economies and societies. Internet of Things (IoT), Artificial intelligence, big data, and high-performance computing technologies will allow increasing levels of autonomy in economic, social and other governance systems. Discovering system parameters, even in the absence of autonomous capabilities, is necessary for decision-making and problem-solving during design and operation.

This paper is part of our broader work on the use of information and communication technology (ICT) to address challenges facing smart cities and societies. Our work on this topic has included the concept of Deep Journalism [137,138], as well as research on topics such as transportation [138], smart families and homes [139], healthcare services for cancer [34], education during the COVID-19 pandemic [140], and AI-based event detection [141]. Future work will be directed to improving the methodological approach presented in this paper using advanced deep learning methods and their applications to investigate and improve labour economics and other problems facing our societies.

**Funding:** "The authors acknowledge with thanks the technical and financial support from the Deanship of Scientific Research (DSR) at the King Abdulaziz University (KAU), Jeddah, Saudi Arabia, under Grant No. RG-11-611-38."

**Acknowledgments:** "The work carried out in this paper is supported by the HPC Center at the King Abdulaziz University."

**Conflicts of Interest:** **"**The authors declare no conflict of interest."